%% file: main.tex
\pdfoutput=1

\documentclass[10pt]{article}
\usepackage[preprint]{tmlr}

\input{math_commands.tex}

\usepackage{hyperref}
\usepackage[utf8]{inputenc}
\usepackage{url}
\usepackage{booktabs}
\usepackage{amsfonts}
\usepackage{nicefrac}
\usepackage{microtype}
\usepackage{xcolor}
\usepackage[many]{tcolorbox}
\usepackage{array}
\usepackage{amsmath}
\usepackage{cleveref}
\usepackage{amsthm}
\usepackage{graphicx}
\usepackage{listings}
\usepackage{xspace}
\usepackage{subcaption}
\usepackage{enumitem}
\usepackage{placeins}
\usepackage{algorithm}

\usepackage{algorithmic}
\usepackage{wrapfig}
\usepackage[T1]{fontenc}
\usepackage{caption}
\usepackage{multirow}
\usepackage{stfloats}
\usepackage{needspace}
\usepackage{tabularx}

\input{cmd_defs}

\crefname{lstlisting}{Listing}{Listings}
\Crefname{lstlisting}{Listing}{Listings}

\title{Interpretable Meta-Policies for Neural Decision Making}

\author{\name Raban Emunds$^{1,2}$ \email raban.emunds@tu-darmstadt.de \\
      \name Jannis Bl\"uml$^{1,2}$ \\
      \name Quentin Delfosse$^{1}$ \\
      \name Kristian Kersting$^{1,2,3,4}$ \\[0.5em]
      \addr $^1$Department of Computer Science, Technical University Darmstadt, Germany \\
      \addr $^2$Hessian Center for Artificial Intelligence (hessian.AI) \\
      \addr $^3$Centre for Cognitive Science, Darmstadt \\
      \addr $^4$German Research Center for Artificial Intelligence (DFKI)}

\begin{document}

\maketitle

\input{sections/abstract}

\input{sections/introduction}

\input{sections/background}

\input{sections/method}

\input{sections/evaluation}

\input{sections/discussion}

\input{sections/related_work}

\input{sections/conclusion}

\bibliography{main}
\bibliographystyle{tmlr}

\newpage
\appendix
\onecolumn
\input{sections/appendix}

\end{document}

%% file: math_commands.tex
\usepackage{amsmath,amsfonts,bm}

\def\eqref#1{equation~\ref{#1}}

\def\1{\bm{1}}

\DeclareMathAlphabet{\mathsfit}{\encodingdefault}{\sfdefault}{m}{sl}
\SetMathAlphabet{\mathsfit}{bold}{\encodingdefault}{\sfdefault}{bx}{n}

%% file: cmd_defs.tex
\newcommand{\method}{NEXUS\xspace}

\newcommand{\neurosymbolic}{neuro-symbolic\xspace}
\newcommand{\Neurosymbolic}{Neuro-symbolic\xspace}

\newcommand{\eg}{e.g.~} 

\newcommand{\ie}{i.e.~} 

\newcommand{\cf}{cf.~}

%% file: sections/abstract.tex
\begin{abstract}
Deep Reinforcement Learning remains difficult to interpret and align due to the black-box nature of neural networks.
While (neuro-)symbolic approaches improve transparency by incorporating symbolic reasoning, applying them directly to low-level actions results in overly complex, unreadable policies.
We introduce \method, a hierarchical RL framework that resolves this tension by combining neural skills with symbolic meta-policies.
By shifting transparent reasoning to high-level skills rather than atomic actions, \method drastically reduces symbolic complexity without sacrificing environment performance.
Object-centric observations enable high-level reasoning and the design of reward functions and meta-policies by interaction with LLMs.
We demonstrate that \method agents learn disentangled skills, are interpretable regarding high-level goals, are better aligned to goals described in the game manual, less prone to reward hacking and more robust to game simplifications than deep RL.
Finally, we isolate and evaluate the impact of different meta-policy architectures (neural, symbolic, \neurosymbolic) on environmental return and aligned task performance. 
\end{abstract}

%% file: sections/introduction.tex
\section{Introduction}
\label{sec:introduction}
\begin{figure}[!bp]
    \centering
    \includegraphics[width=\textwidth]{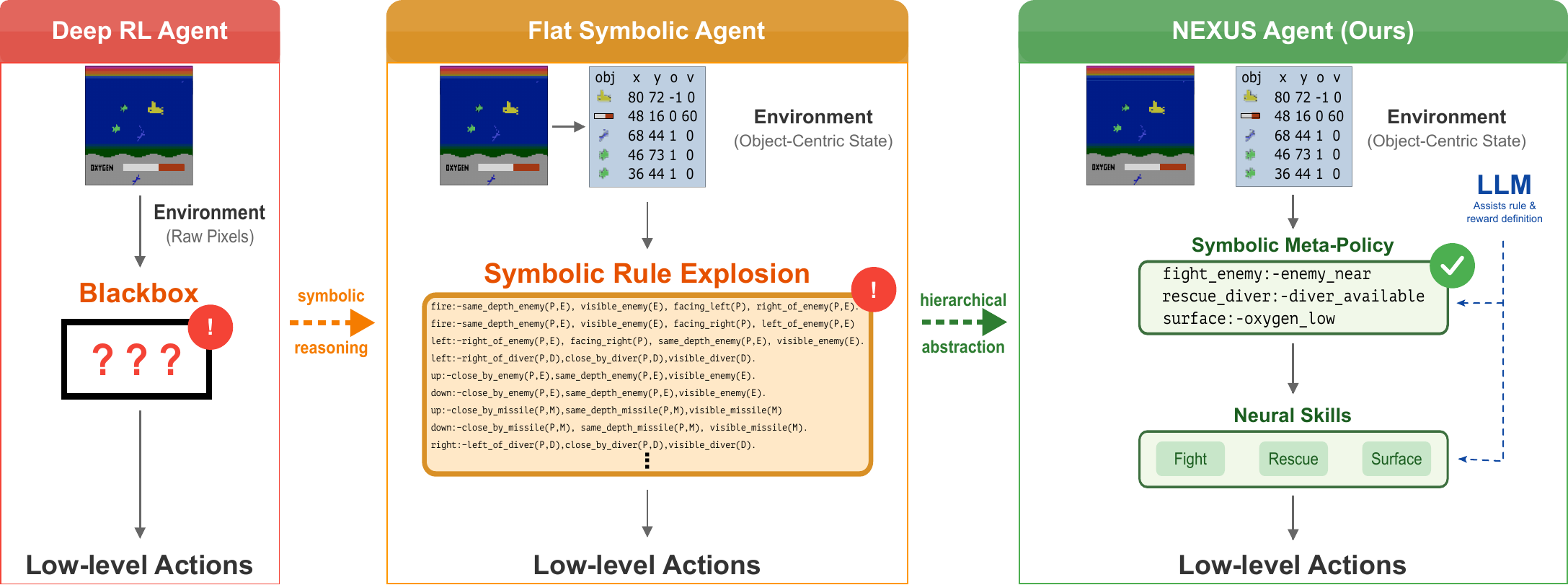}
    \caption{ \textbf{NEXUS provides high-level interpretability through symbolic reasoning and abstraction.}
    Neural network-based policies are non-modular black box models. While symbolic approaches are fully transparent by leveraging logical rules or decision trees, they quickly become excessively complex if applied to low-level actions. NEXUS abstracts low-level actions into skills and provides transparent high-level decisions.} 
    \label{fig:nexus_fig1}
\end{figure}

Deep reinforcement learning (DRL) policies are highly capable \citep{mnih2015humanlevel, schulman2017proximal, gallici2024simplifying} but operate as black boxes, exhibiting behaviors that are difficult to interpret and potentially misaligned \citep{rudin2019stop}. This opacity makes identifying and correcting undesirable behaviors a persistent challenge for practitioners \citep{zahavy2016graying, zhang2018study, delfosse2024interpretable}.

\Neurosymbolic approaches address this issue by combining neural networks for perceptual grounding with symbolic reasoning modules for decision-making \citep{delfosse2023interpretable, hazra2023deep, acharya2024neurosymbolic, delfosse2024interpretable, luo2024endtoend}.
These architectures aim to enhance transparency by representing the policies through symbolic structures that are more readily interpretable.
Despite their promise, applying symbolic reasoning directly to low-level action spaces often results in policies of prohibitive complexity (\cf~\Cref{fig:nexus_fig1} center, Appendix~\ref{apdx:motiv}).
The combinatorial explosion of symbolic rules at fine-grained action levels undermines interpretability and scalability.

Hierarchical Reinforcement Learning offers an alternative by abstracting sequences of actions into higher-level skills or options \citep{sutton1999between, dietterich2000hierarchical, barto2003recent}.
While the framework provides a natural structure, the resulting options, when learned autonomously, are often entangled or have overlapping goals, and the meta-policy remains opaque, both hindering interpretability.

To address these limitations, we draw inspiration from the dual-process theory of cognition \citep{kahneman2011thinking}, where fast, intuitive processes and slower, deliberative reasoning continuously interact. We emulate this interplay in our hierarchical framework \method, \textbf{Ne}ural \textbf{EX}ecution \textbf{U}nder \textbf{S}ymbols: a low-level neural policy provides efficient, automatic execution (analogous to System 1), while a meta-policy layer of simple, interpretable rule sets (analogous to System 2) constantly guides skill selection, preserving clarity about the reasoning behind the active skill in complex environments (\cf~\Cref{fig:nexus_fig1}, right). We ensure disentangled skills, \ie each skill corresponds to a distinct situation with clear semantics, by defining their subgoals explicitly through reward functions on objects extracted from the image.

To reduce the manual effort required, we leverage the reasoning capabilities of Large Language Models (LLMs) to assist in identifying those skills, defining the high-level meta-policy functions, and generating the reward signals that guide the training of the low-level neural controllers.
Both the low-level skills and the meta-policy Q-function are learned jointly in an off-policy manner.
Overall, our \neurosymbolic design enables 
inspectable high-level policies
without compromising the efficiency of neural agents.

Our primary contributions are as follows:
\begin{description}[leftmargin=20pt, itemsep=0pt,parsep=0pt,topsep=-3pt,partopsep=0pt]
\item[(i)] We extend \emph{Parallelised Q-Networks} to the hierarchical setting for efficient and scalable meta-policy and skill learning (\Cref{method:HPQN}).
\item[(ii)] We introduce \method and compare three meta-policy variants with varying degrees of interpretability and flexibility (\Cref{method:HPQN} - \Cref{method:soft}).
\item[(iii)] We demonstrate that object-centric rewards and high-level meta-policy functions guide training towards disentangled skills and an interpretable high-level skill selection (\Cref{exp:interpret}).
\item[(iv)] We show that \method agents are competitive to DRL methods while being less susceptible to reward hacking of environment rewards and more robust to game simplifications. (\Cref{exp:performance}).
\end{description}

%% file: sections/background.tex
\section{Background}
\label{sec:background}
\method builds upon the deep Reinforcement Learning method \emph{Parallelised Q-Networks}, which is a \emph{Deep Q-Networks} variant that scales favorably and \emph{Hierarchical Reinforcement Learning} that decomposes complex tasks into hierarchies of simpler sub-tasks, thereby abstracting actions into skills.

\paragraph{Deep and Parallelised Q-Networks.} 
As sequential decision making framework we adopt Reinforcement Learning (RL), in which an agent learns to interact with an environment in order to maximize cumulative reward.
The environment is typically modeled as a Markov Decision Process (MDP), defined by the tuple \( \mathcal{M} = \langle\mathcal{S}, \mathcal{A}, P, R, \gamma \rangle \), where \( \mathcal{S} \) is the set of states, \( \mathcal{A} \) is the set of actions, \( P(s' \mid s, a) \) is the transition probability from state \( s \) to state \( s' \) under action \( a \), \( R(s, a) \) is the reward function and \( \gamma \in [0,1) \) is the discount factor.
The goal of the agent is to learn a policy \( \pi: \mathcal{S} \to \mathcal{A} \) that maximizes the expected discounted return: $\mathbb{E}_\pi \left[ \sum_{t=0}^{\infty} \gamma^t R(s_t, a_t) \right]$.
In \textit{Q-learning}, a value-based RL algorithm, the agent seeks to learn the optimal action-value function \( Q^*(s, a) \), which satisfies the Bellman optimality equation:
$Q^*(s, a) = \mathbb{E}_{s'} \left[ R(s, a) + \gamma \max_{a'} Q^*(s', a') \right]$.
This function is updated iteratively via the Q-learning update rule: $Q(s_t, a_t) \leftarrow Q(s_t, a_t) + \alpha \left[ r_t + \gamma \max_{a'} Q(s_{t+1}, a') - Q(s_t, a_t) \right]$,
where \( \alpha \) is the learning rate.
With growing state space \( \mathcal{S} \), it becomes infeasible to store and update a tabular \( Q \)-function. 
\textit{Deep Q-Networks (DQN)} \citep{mnih2015humanlevel} address this by using a deep neural network \( Q_\theta(s, a) \), parameterized by \( \theta \), to approximate \( Q(s, a) \).
DQN introduces several key modifications to stabilize learning, including experience replay, where transitions \( (s_t, a_t, r_t, s_{t+1}) \) are stored in a replay buffer and sampled randomly to break correlations between consecutive updates and target networks, where a separate network \( Q_{\theta^-}(s, a) \) is used to compute the target value and is updated periodically: $y_t = r_t + \gamma \max_{a'} Q_{\theta^-}(s_{t+1}, a').$%

More recently, \citet{gallici2024simplifying} introduced \textit{Parallelised Q-Networks (PQN)}, a simplified variant of DQN that eliminates the use of experience replay and target networks.
Instead, PQN leverages a large number of parallel (ideally vectorized) environments and applies normalization techniques to mitigate training instabilities.
This high degree of parallelization enables substantially faster training while maintaining performance.

\paragraph{Hierarchical Reinforcement Learning.}
\emph{Hierarchical reinforcement learning (HRL)} decomposes complex tasks into simpler sub-tasks, often modeled as temporally extended actions or \emph{options} \citep{sutton1999between}. 
At inference time, a meta policy selects the option, enabling more efficient exploration and transfer in environments with long horizons or sparse rewards. 
Sub-policies can be learned autonomously using intrinsic objectives \citep{bacon2017optioncritic, vezhnevets2017feudal} or be guided by manually designed reward functions \citep{sutton1999between, dietterich2000hierarchical}. 
However, autonomous discovery often suffers from option entanglement or collapse, where multiple options overlap or only a single option is actively used \citep{harb2017when, hyun2019disentangling}.
\method instead enforces its skills to be disentangled by explicitly defining reward functions for each skill, which can be LLM-generated.

%% file: sections/method.tex
\section{Neural Execution Under Symbols}
\label{sec:method}

\begin{figure*}[tb]
    \centering
    \includegraphics[width=\textwidth]{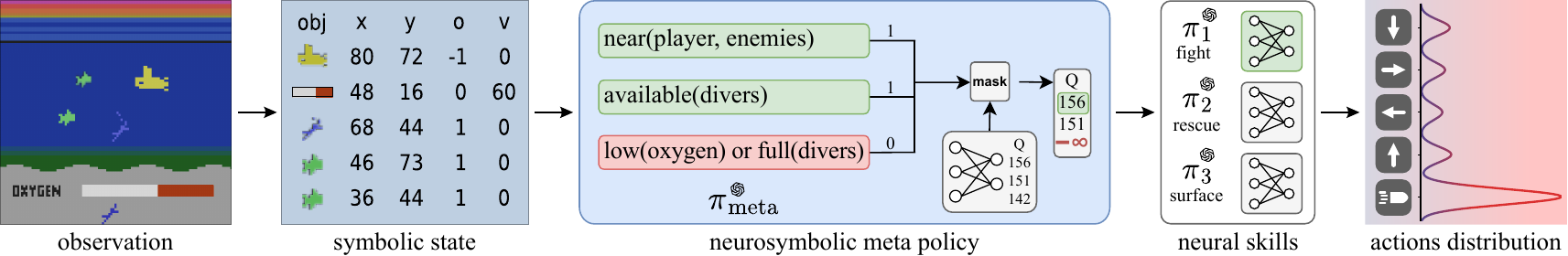}
    \caption{\textbf{\method with a \neurosymbolic meta-policy.} First, the symbolic state is extracted from the observation. 
    Then, the LLM-generated rules filter potential relevant skills, which the trained meta-policy Q-function selects between.
    Finally, the selected neural skill outputs the low-level action.
    Using the same experience, the meta-policy's Q-function is trained to optimize the environmental reward, while each skill is trained using its own LLM-generated reward, yielding disentangled skills with interpretable objectives.}
    \label{fig:pipeline}
\end{figure*}

To bridge the gap between well performing deep neural RL and human-interpretable logic, we introduce \method, a hierarchical RL framework that integrates symbolic reasoning with neural skill execution.
The overall pipeline is described in \Cref{fig:pipeline}.
At a high level, \method decomposes decision-making into two layers: (1) a high-level meta-policy and (2) a set of low-level neural skill policies optimized for specific sub-goals.
At each step, the decision process proceeds as follows.
First, object-centric representations are extracted from pixel-based inputs.
Second, the meta-policy selects the most suitable skill for the current situation with the goal of maximizing environmental reward.
Finally, the chosen skill executes its low-level policy to maximize the corresponding sub-goal.
Object-centric representations are crucial for enabling symbolic rule definition and skill-specific reward design, both of which are difficult to obtain without such structured representations.
Moreover, they allow LLMs to generate those rules and reward functions.

In the following, we describe, three viable types of meta-policy function: purely neural, purely symbolic, and \neurosymbolic.
The first method \method (neural), \cf~\Cref{method:HPQN}, corresponds to a hierarchical PQN method, which learns both a neural meta-policy and neural skills (based on skill-specific rewards) and which serves as the foundation for subsequent variants. 
While it offers disentangled skills and high flexibility due to the learned meta-policy, it is missing the ability to reason about why a given sub-goal (and the corresponding skill) is selected in the current step.
This shortcoming can be facilitated by replacing the learned meta-policy with a fixed, interpretable function which can either be manually defined or LLM-generated.
This variant, \method (symbolic), \cf~\Cref{method:fixed}, is fully symbolic at the meta-policy level and thus produces the easiest to interpret meta-policy of the investigated options.
Finally, we explore a hybrid approach \method (nesy), \cf~\Cref{method:soft}, which filters candidate skills via predefined symbolic rules, but then selects the final skill using Q-learning on the abstract skill level in ambiguous situations.

\subsection{Hierarchical PQN}
\label{method:HPQN}
We accommodate the hierarchical learning setup by modeling the environment as two layers of MDPs.
On the action level, we define a collection of MDPs \( \bm{\mathcal{M}} = \{ \langle\mathcal{S}, \mathcal{A}, P, R_n, \gamma \rangle \}_{n=1}^N \) for \(N\) skills, each associated with a distinct option policy \(\pi_n \in \Pi\) that decides over the actual environment actions.
The meta-level MDP \( \mathcal{M}_\text{meta} = \langle \mathcal{S}, \Pi, P, R_\text{env}, \gamma \rangle \) governs option selection through a meta-policy \(\pi_\text{meta}\) that selects over the options \(\Pi\).
Note that the option-level MDPs differ only in their reward functions \(R_n\), while the meta-level MDP retains the environment’s original reward \(R_\text{env}\).

To enable learning across sub-policies even when they are not actively selected, we adopt off-policy Q-learning. This allows all the skills to learn from trajectories generated by the active skill and meta-policy while optimizing for their respective reward structures. 
Exploration within each skill is conducted via $\epsilon$-greedy action selection.
For scalability through vectorization, we build our hierarchical training approach upon PQL with $\lambda$-returns \citep{gallici2024simplifying}.
However, instead of learning a single shared Q-network $Q_\phi$ as in PQL, we learn separate Q-functions $Q_{\phi_n}$ for each low-level skill, along with a meta-level Q-network $Q_{\phi_\text{meta}}$.

For updating, we roll out trajectories $(s_0, ..., s_T)$ following the global policy by selecting the next active skill $\pi_n \in \Pi$ using the meta-policy Q-function $Q_{\phi_\text{meta}}$ and find the most promising action according to the active skill's Q-function $Q_{\phi_n}$:
\begin{equation}
\small
\pi_n = \arg \max_{\pi_n \in \Pi} Q_{\phi_{\text{meta}}}(s_t, \pi_n), \quad a' = \arg \max_{a' \in \mathcal{A}} Q_{\phi_n}(s_t, a').
\end{equation}
By executing the action in the environment, we obtain the rewards and the next environmental state for the next iteration.
We employ $\epsilon$-greedy for exploration during both the skill and the action selection.

Next to the environment reward $r_{\text{env}, t}$, we require skill-specific reward functions $r_{n, t}$ that are based on the object-centric state $s_t$ and may be automatically generated by LLMs (\cf~Appendix~\ref{sec:llm_interaction}), which we use to compute the $\lambda$-returns recursively back in time (for details, we refer the reader to \citet{gallici2024simplifying, daley2019reconciling}):
\begin{equation}
R_{n,t}^\lambda = r_{n,t} + \gamma \left[ \lambda R_{n,{\scriptscriptstyle t+1}}^\lambda + (1-\lambda) \max_{a'} Q_{\phi_n}(s_{\scriptscriptstyle t+1}, a') \right]\!, 
\end{equation}
and similarly, for the learned meta-policy using environment rewards: 
\begin{equation}
R_{\text{env},t}^\lambda =\! { r_{\text{env},t}\! +\! \gamma \!\left[ \lambda R_{\text{env},{\scriptscriptstyle t+1}}^\lambda\! +\! (1\!-\!\lambda)\!\max_{\pi_n} Q_{\phi_\text{meta}}(s_{\scriptscriptstyle t+1}, \pi_{\scriptscriptstyle n}) \right]},
\end{equation}
or, if $s_t$ is terminal $R_{n,t}^\lambda = r_{n,t}$ and $R_{\text{env},t}^\lambda = r_{\text{env},t}$.
All learned Q-functions are updated towards their $\lambda$-returns.
We provide the full algorithm in Appendix~\ref{apdx:hpqn}.

\subsection{Interpretable Meta-Policy Function}
\label{method:fixed}
While the hierarchical structure allows to identify the active skill, it does not expose the decision process that leads to its selection.
To make this process transparent, we introduce interpretable meta-policy functions implemented as rule-based programs.
We choose this representation for two key reasons.
First, rule-based programs offer simplicity and accessibility since conditional statements can typically be understood and modified even by users with limited programming background.
Second, LLMs are highly effective at generating and editing code-like structures due to extensive pretraining on programming data.
This enables generation of human-interpretable meta-policies and greatly reduces manual effort.
Further details on how we employ LLMs to generate the meta-policy function are available in Appendix~\ref{sec:llm_interaction}.

Formally, instead of relying on a learned Q-function \( Q_{\phi_\text{meta}} \), we define a meta-policy \( \pi_{\text{meta}} : \mathcal{S} \rightarrow \Pi \) as a set of human-readable rules that directly maps the current object-centric state \( s_t \in \mathcal{S} \) to the selected low-level policy \( \pi_n \in \Pi \):
\begin{equation}
\pi_n = \pi_{\text{meta}}(s_t).
\end{equation}

We hereby constrain \( \pi_{\text{meta}} \) to a set of rules \( \{ (A_k, \pi_{(k)}) \} \), where each condition set \( A_k \subseteq \mathcal{S} \) defines the state subspace where policy \( \pi_{(k)} \) is triggered. To enable transparent inspection of policy decisions, we require these activation conditions to be pairwise disjoint (\ie, \( A_i \cap A_j = \emptyset \) for all \( i \neq j \)). This mutual exclusivity ensures that the selected skill can always be unambiguously traced back to the single unique rule that evaluated to true given the current state. Example rules are available in the evaluation section (\cf~\Cref{lst:symbolic_meta_pol}).

\subsection{\Neurosymbolic Meta-Policy Function}
\label{method:soft}

The symbolic meta-policy is a deterministic function that selects mutually exclusive skills based on the rule with the highest priority.
This approach may become ineffective in situations where multiple skills are simultaneously applicable.
Consider an agent controlling a submarine that can shoot a nearby enemy or surface to replenish oxygen.
The most viable rule depends on the specific context: if the oxygen level is critically low, resurfacing may be prioritized; if an enemy is dangerously close, attacking may take precedence.
Crafting hard-coded rules to handle all these situations, could lead to growing, less interpretable policies.

\begin{wrapfigure}{r}{0.6\textwidth}
\vspace{-8pt}
\centering
\begin{minipage}{\linewidth}
\begin{algorithm}[H]
\caption{\method Variants}
\label{alg:h_pqn_var}
\begin{algorithmic}[1]
\REQUIRE Skill policies $\{\pi_n\}_{n=1}^N$ with Q-functions $\{Q_{\phi_n}\}_{n=1}^N$, meta-policy $Q_{\phi_{\text{meta}}}$ (for learned/soft)
\FOR{each episode}
    \FOR{each environment step (in parallel)}
        \STATE Extract object-centric state $s_t$ 
        \STATE Select skill $\pi_{n,t}$ as follows:\\
        A) Neural: $\pi_{n,t} = \arg\max_{\pi_n} Q_{\phi_{\text{meta}}}(s_t, \pi_n)$\\
        B) Symbolic: $\pi_{n,t} = \pi_{\text{meta}}(s_t)$\\
        C) NeSy: $\pi_{n,t}\! =\! \arg\max_{\pi_n:\, c_{t,n}=1} Q_{\phi_{\text{meta}}}(s_t, \pi_n)$\\
        \STATE Select action $a_t$ using $\epsilon$-greedy from $\pi_{n,t}$
        \STATE Execute $a_t$ and observe transition $(\{r_{n,t}\}_{n=0}^N, r_{\text{env},t}, s_{t+1})$
    \ENDFOR
    \FOR{each gradient step}
        \STATE Compute $R_{n,t}^\lambda$ and $R_{\text{env},t}^\lambda$ (if A or C)
        \STATE Update each $Q_{\phi_n}$ using skill-specific $R_{n,t}^\lambda$
        \STATE Update $Q_{\phi_{\text{meta}}}$ using $R_{\text{env},t}^\lambda$ (if A or C)
    \ENDFOR
\ENDFOR
\end{algorithmic}
\end{algorithm}
\end{minipage}
\end{wrapfigure}

As an alternative meta-policy variant aimed at automatically resolving such situations, we also explore a \neurosymbolic (NeSy) approach. Similar to the symbolic meta-policy, this variant maintains a set of high-level, interpretable conditions (e.g., ``go to surface if oxygen is low''), represented as a binary condition vector $c_t \in \{0,1\}^N$ indicating rule activation at time $t$. These rule-sets are again formulated as simple code functions derived from the symbolic meta-policies; however, in this variant, we allow multiple conditions to be active concurrently. The skill selection is then restricted by these conditions: the meta-policy selects the skill with the highest Q-value among those whose condition is active:
\begin{equation}
\begin{aligned}
    \pi_n &= \arg\max_{\pi_n \in \Pi_t} Q_{\phi_{meta}}(s_t, \pi_n), \\
    &\text{where } \Pi_t = \{\pi_n \in \Pi \mid c_{t,n} = 1\}.
\end{aligned}
\end{equation}
where $\Pi_t \subseteq \Pi$ denotes the set of skills whose conditions are active at
time $t$. In practice, this corresponds to a masked argmax in which the Q-values of
inactive skills are set to $-\infty$.
This design successfully maintains interpretability at the symbolic level and simultaneously enables the agent to resolve ambiguous scenarios adaptively based on learned preferences. An overview over the entire NEXUS algorithm with the different meta-policy variations is provided in Algorithm \ref{alg:h_pqn_var}.

%% file: sections/evaluation.tex
\section{Experimental Evaluation}

\label{sec:exps}
This work aims to develop an object-centric pipeline for agents capable of solving complex environments while maintaining interpretability of high-level goals. We assess this objective by addressing the following research questions:
\begin{description}[leftmargin=20pt, itemsep=0pt,parsep=0pt,topsep=-3pt,partopsep=0pt]
\item[(Q1)] Does \method learn meaningful disentangled skills?
\item[(Q2)]  Is the high-level skill selection of NEXUS interpretable?
\item[(Q3)] Can \method compete with other deep methods?
\item[(Q4)] Does \method improve robustness to game simplifications?
\end{description}

\begin{figure*}[tb]
    \centering
    \includegraphics[width=\linewidth]{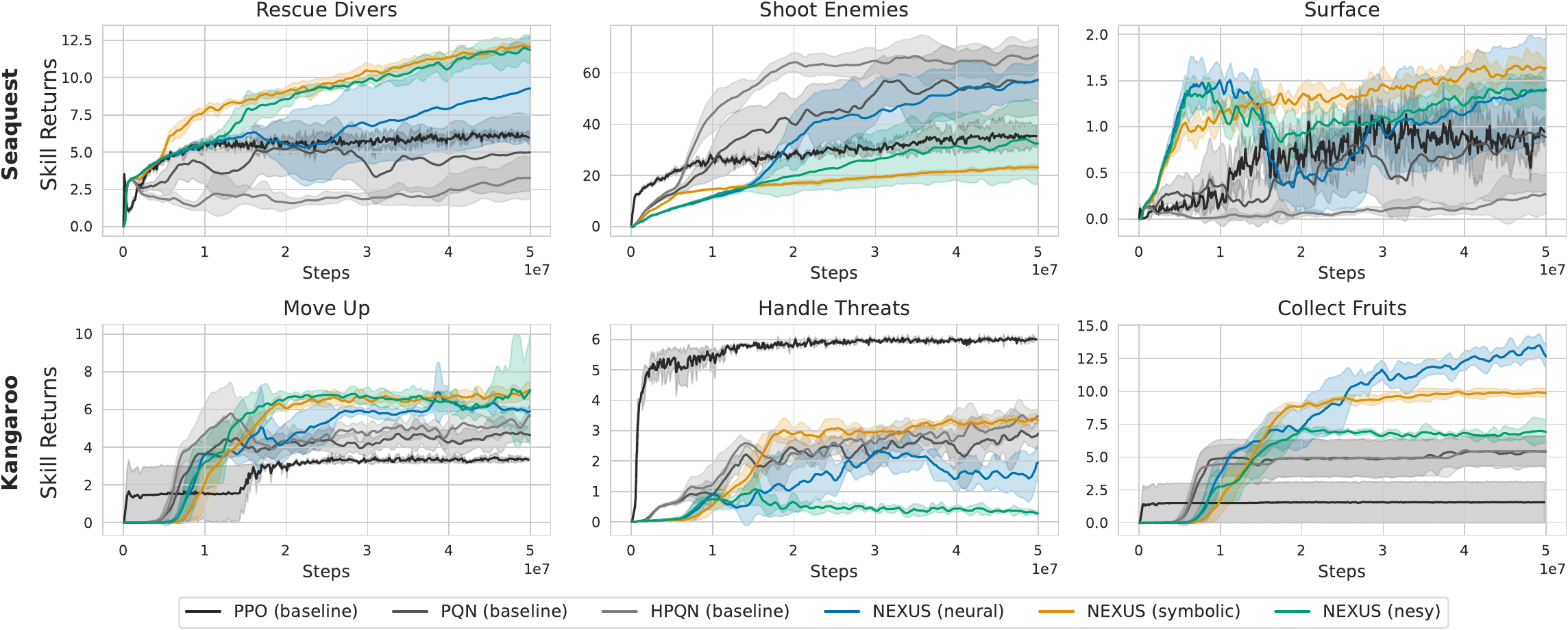}
    \caption{\textbf{\method learns disentangled skills from off-policy data.} In Seaquest (top), the baseline methods mainly focus on shooting enemies, while the \method approaches acquire the target skills more evenly.
    Similarly, in Kangaroo (bottom), the \method approaches learn 'Move Up' and 'Collect Fruits' reliably, while the baselines focus mostly on 'Handle Threats'.
    Aggregated over 6 seeds, $\pm 1$ standard deviation.}
    \label{fig:skill_learn}
\end{figure*}

\begin{figure*}[b]
    \centering
    \includegraphics[width=\linewidth]{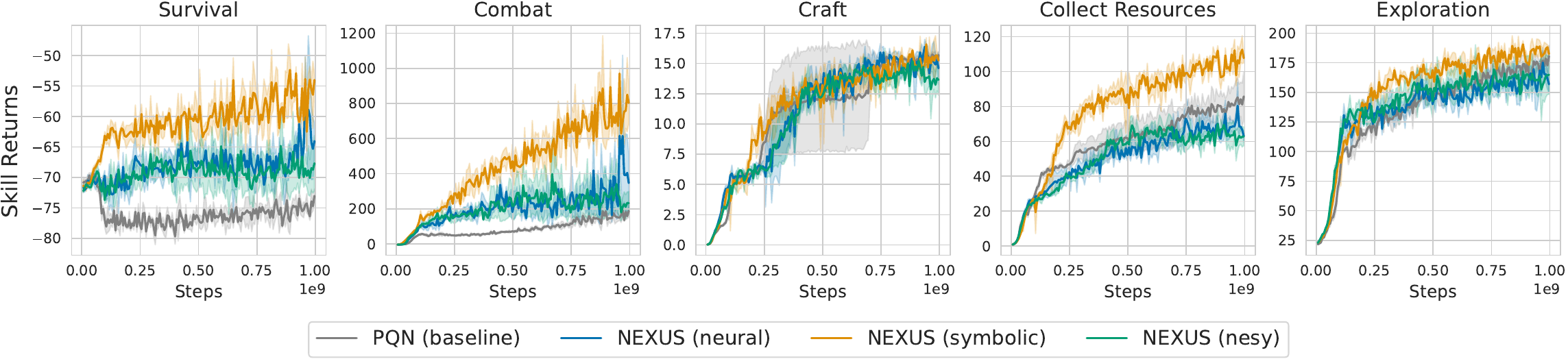}
    \caption{\textbf{\method succesfully learns LLM-proposed Crafter skills.} Compared to the baseline, \method approaches often converge faster on the proposed skills. Aggregated over 6 seeds, $\pm 1$ standard deviation.}
    \label{fig:crafter}
\end{figure*}

\subsection{Experimental Setup}
Our experiments assume access to pre-extracted object representations and attributes. We therefore provide the agent with symbolic environment states directly, allowing us to remain agnostic to the specific object extraction pipeline.
We conduct experiments on \emph{JAXAtari}~\citep{jaxatari2026}, a JAX-based reimplementation of the Atari Learning Environment (ALE) \citep{bellemare2013arcade}, specifically using the games \textit{Kangaroo} and \textit{Seaquest}, with object-centric representations similar to \textit{OCatari} \citep{delfosse2024ocatari}.
Additionally, we evaluate on \textit{Crafter} \citep{hafner2021crafter}, using the JAX reimplementation by \citet{matthews2024craftax}, an environment that integrates elements from the games Minecraft and NetHack (\cf~Appendix~\ref{apdx:envs}).
Finally, for generalization we also test the trained agents on slightly modified versions of the games.
These modifications are designed to highlight the sensitivity of standard reinforcement learning agents to game simplifications.
For example, the simplification of removing enemies (i.e., sharks and submarines) in Seaquest already leads to a substantial drop in many deep agents performances~\citep{delfosse2025deep}.

We compare the three developed meta-policy variations of \method to the default, non-interpretable PQN \citep{gallici2024simplifying}, and, in case of the Atari games, with the actor-critic PPO \citep{schulman2017proximal} as baselines.
Additionally, we compare to HPQN, a hierarchical PQN variation that does not employ skill-specific rewards and has a purely neural meta-policy.
All results (including baselines) are directly trained on object-centric inputs, which may affect performance compared to image based training.
We adhere to the standard frame budgets of both games: 200M frames for JAXAtari and 1B frames for Crafter (\cf~Appendix~\ref{apdx:Impl}).
All metrics shown are aggregated over a total of six seeds.

\subsection{Interpretability Results}
\label{exp:interpret}

\paragraph{Disentangled skill learning (Q1).} 
We first evaluate whether \method enables efficient learning of meaningful skills.
Since each skill requires its own reward function, we leverage the reasoning capabilities of LLMs to generate them based on the game manual, skill definitions, and the object-centric state (\cf Appendix~\ref{sec:llm_interaction}).

\Cref{fig:skill_learn} demonstrates that \method is able to learn most target skills successfully from a single source of off-policy data.
This confirms that a skill can learn from another skills action and is not required to be activated to do so. 
Unlike the baselines, which tend to mostly focus on a single skill that maximizes environment reward (e.g., Shoot Enemies for Seaquest, Handle Threats for Kangaroo), \method approaches promote balanced skill acquisition.
We experience a similar picture when evaluating on Crafter (\cf~\Cref{fig:crafter}).
Using the game manual and state description, we query an LLM for essential skills. Recurrent versions of \method are then trained explicitly for these skills and compared to a recurrent PQN baseline. 
\method approaches are able to learn all five skills simultaneously, while the PQN baseline focuses only on some skills.
An additional baseline comparison to PQN with shaped rewards is available in Appendix~\ref{apdx:shaped_rewards}.

\paragraph{Interpretability of \method meta-policies (Q2).}
\begin{figure*}[tb]
    \centering
    \includegraphics[width=\linewidth]{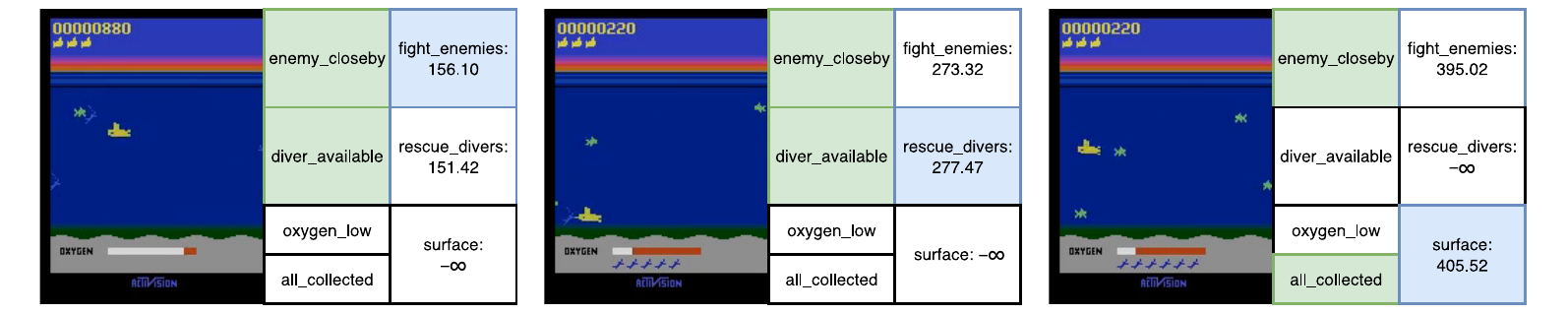}
    \caption{\textbf{\method (nesy) restricts skill selection through interpretable rules but remains flexible in ambiguous scenarios.}
    Left: \texttt{enemy\_closeby} and \texttt{diver\_available} both hold; the meta-policy prefers fighting, likely as the diver is near the enemy. Center: with the diver more accessible, rescuing is preferred. Right: with all divers collected and an enemy nearby, surfacing has the higher estimated return.}%
    \label{fig:interpretable}
\end{figure*}

\begin{figure}[tb]
  \centering
  \begin{minipage}[t]{0.48\textwidth}
    \begin{lstlisting}[
      language=Python, 
      caption={Symbolic meta-policy for Seaquest.}, 
      label={lst:symbolic_meta_pol},
      basicstyle=\scriptsize\ttfamily,
      breaklines=true
    ]
def meta_policy(st: State):
    if oxygen_low(st) or \
         all_collected(st.divers):
        return surface
    elif enemy_closeby(st):
        return fight_enemies
    if diver_available(st):
        return rescue_divers
    return fight_enemies
    \end{lstlisting}
  \end{minipage}%
  \hfill
  \begin{minipage}[t]{0.48\textwidth}
    \begin{lstlisting}[
      language=Python, 
      caption={Neurosymbolic meta-policy rules Seaquest.}, 
      label={lst:nesy_meta_pol},
      basicstyle=\scriptsize\ttfamily,
      breaklines=true
    ]
def meta_policy_rules(st: State):
    fight_enemies, rescue_divers, surface = \
        True, False, False
    if diver_available(st):
        rescue_divers = True
    if oxygen_low(st) or \
       all_collected(st.divers):
        surface = True
    return [fight_enemies, rescue_divers, surface]
    \end{lstlisting}
  \end{minipage}
\end{figure}

Next, we assess the interpretability of \method by inspecting a fixed meta-policy for Seaquest, shown in \Cref{lst:symbolic_meta_pol}.
By abstracting raw observations into object-centric representations and low-level actions into high-level skills, the high-level decision-making process becomes transparent, while the execution of each skill remains neural. Each option has a clear, mutually exclusive activation condition, enabling unambiguous skill selection. 
For example, the surfacing skill takes precedence whenever oxygen is low or all divers are collected.
The resulting meta-policy is fully transparent on the high-level decisions and easily modifiable by humans.
Additionally, the simplicity in rule design allows for both manual design by technical users and automated generation by LLMs~(\cf Appendix~\ref{sec:llm_interaction}). 
For neuro-symbolic meta-policies, multiple conditions may be satisfied simultaneously (e.g., an enemy is nearby and oxygen is low; \cf~\Cref{lst:nesy_meta_pol}), which occurs frequently in practice (\cf~\Cref{apdx:nesy_analysis}). In these cases, the rules transparently restrict the set of admissible skills, but the final choice among them is made by the learned meta-policy Q-function and is therefore only partially interpretable. The neuro-symbolic variant thus trades interpretability for flexibility.
\Cref{fig:interpretable} visualizes the \neurosymbolic meta-policy controlling a Seaquest agent while maintaining transparency regarding the current goals and decision logic. We analyze the neuro-symbolic meta-policy and skill activation frequencies in Appendix \ref{apdx:nesy_analysis}.

\subsubsection*{Policy complexity}

To quantify the interpretability of the high-level decision logic without human studies, we measure its simulatability \citep{lipton2018mythos} with two metrics: Average Decision Path Length (ADPL) \citep{wu2017beyond}, the expected number of condition checks per decision, and Rule Size / Length \citep{lakkaraju2016interpretable}, the number of rules and their total length. Both cover only the symbolic layer, not the neural skills.

Compared to the flat interpretable baseline NUDGE~\citep{delfosse2023interpretable} on Seaquest, \method reduces policy complexity by almost an order of magnitude (\cf \Cref{tab:pol_compl}, Appendix~\ref{apdx:interpret_measures}): NUDGE requires 16 rules with 46 literals and an ADPL of 23.56, whereas delegating low-level control to neural skills leaves \method with 2--4 rules, 3--4 predicates and an ADPL of 2.5--3, well within human working-memory bounds ($7 \pm 2$ concepts; \citealp{miller1956magical, lage2019evaluation}).

\begin{table}[tb]
\centering
\caption{Quantitative interpretability metrics of the symbolic decision logic, comparing NUDGE (action level) and NEXUS (skill level) on Seaquest. NEXUS metrics cover the meta-policy only; low-level skills are neural.}
\label{tab:pol_compl}
\begin{tabular}{lcc}
\toprule
\textbf{Policy} & \textbf{Rule Size and Length} \cite{lakkaraju2016interpretable} & \textbf{ADPL} \cite{wu2017beyond} \\
\midrule
NUDGE (\Cref{lst:nudge}) & 16 rules, 46 predicates & 23.56 evaluations \\
\method (\Cref{lst:symbolic_meta_pol}) & 4 rules, 4 predicates & 3 evaluations \\
NEXUS (\Cref{lst:nesy_meta_pol}) & 2 rules, 3 predicates & 2.50 evaluations \\
\bottomrule
\end{tabular}
\end{table}

\subsection{Performance Results}
\label{exp:performance}
\paragraph{Comparison to other approaches (Q3).}
\begin{wrapfigure}{r}{0.6\textwidth}
  \centering
  \vspace{-20pt}
  \includegraphics[width=\linewidth]{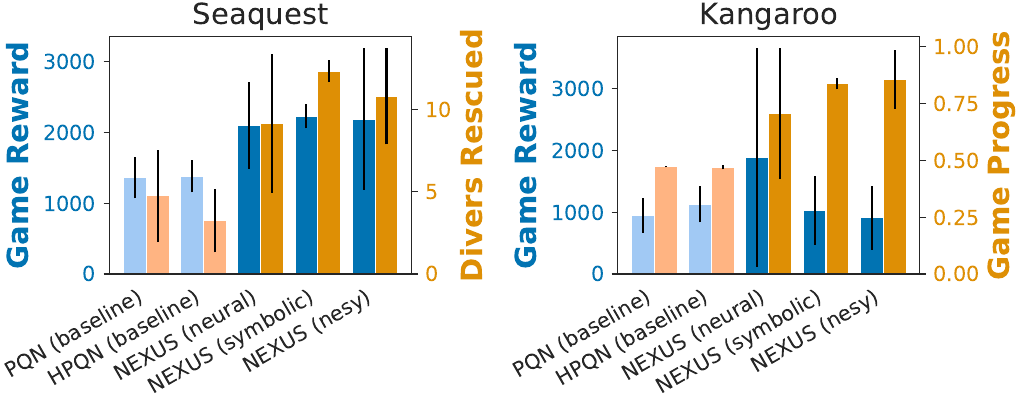}
  \caption{\textbf{\method is competitive.} In Seaquest, \method outperforms the baselines on both game reward and rescuing divers. In Kangaroo, \method approaches progress farther. Aggregated over 6 seeds, $\pm 1$ standard deviation.}
  \label{fig:returns}
  \vspace{-10pt}
\end{wrapfigure}
We evaluate \method against baselines using two metrics: Game returns and aligned environment goals, which track progress toward goals defined in game manuals (\eg, divers rescued in Seaquest, level completion in Kangaroo).
The latter aims to capture the overall alignment with the intended game objectives, which differ from simply maximizing the environmental reward, a phenomenon in deep RL known as reward hacking.
As shown in \Cref{fig:returns}, \method approaches are competitive to the baselines.
In most cases, they achieve comparable or even higher game reward and notably outperform the baselines on the games' actual goal described in the manual, demonstrating the baselines reward hacking tendency.

The results suggest that \method mitigates reward misalignment by incorporating domain priors (game manual) into the decision-making process, while still achieving good overall game performance.
We provide additional evaluations on more environments and compare \method to other \neurosymbolic RL methods in Appendix~\ref{apdx:add_results}.

\begin{figure*}[tb]
    \centering
    \includegraphics[width=\linewidth]{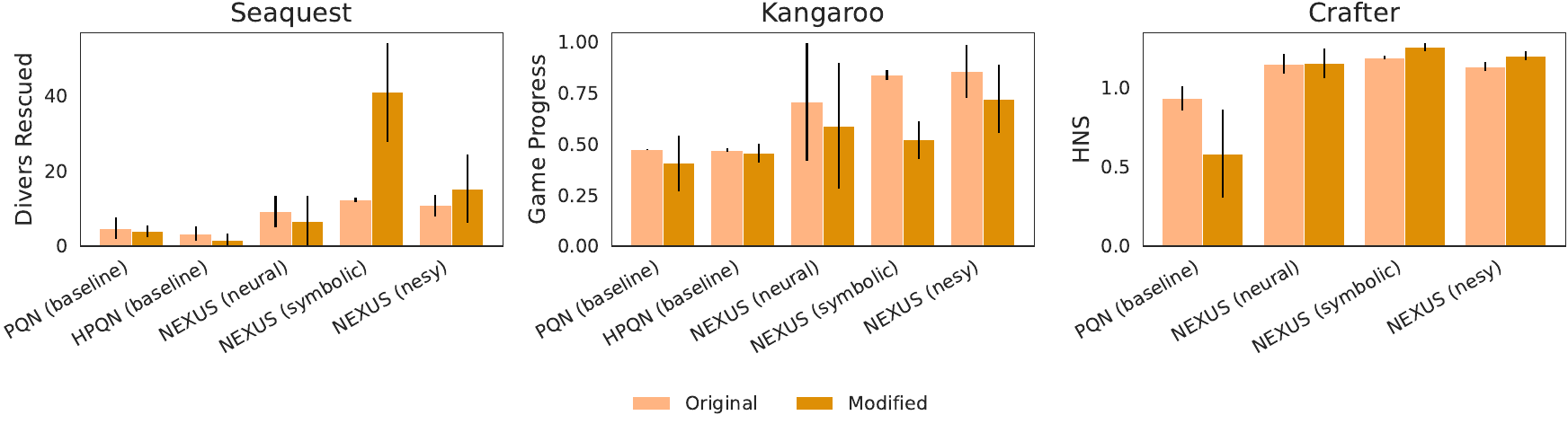}
    \caption{\textbf{\method remains practical on simplified games.} While baseline performance on aligned metrics remains low, \neurosymbolic and fully symbolic \method improve on modified Seaquest (left). Modified kangaroo (middle) reduces performance of symbolic \method, while \neurosymbolic \method remains effective. Crafter performance (right) remains consistent or improves for all NEXUS versions. Aggregated over 6 seeds, $\pm 1$ standard deviation.} 
    \label{fig:dist-shift}
\end{figure*}

\paragraph{Robustness to game simplifications (Q4).}
Most RL algorithms struggle to adapt to even minor variations in the environment \citep{delfosse2025deep}.
Surprisingly, their performances drop even in settings that simplify the game for humans, such as removing deadly threats like enemies and their projectiles in Seaquest and Kangaroo.
We assess whether \method agents can generalize to such simplified variants by training the agents on the standard versions of the games and evaluating on the unseen simplifications.
In Seaquest and Kangaroo we remove all threats, while in Crafter we remove the need to drink water for survival.
\Cref{fig:dist-shift} presents the results.
As expected, the baselines suffer performance degradation under the simplifications or already had low scores.
In contrast, the \method variants display varying degrees of robustness.
In Atari, the symbolic and \neurosymbolic variants demonstrate improved robustness, with small performance drops in Kangaroo and even gains in Seaquest, while the fully neural variant shows limited robustness.
In Crafter, all \method variants maintain or improve performance.
We attribute this robustness to symbolic steering of the meta-policy, which deactivates unnecessary skills (handling threats, finding water).

%% file: sections/discussion.tex
\section{Discussion}
Overall, the empirical results highlight the strengths of our approach across multiple dimensions. 
Compared to the DRL baselines, \method achieves competitive game rewards and exhibits a strong advantage when evaluated on metrics measuring game progress.
Furthermore, we observe improved robustness against environmental simplifications.
Taken together, these findings strongly suggest that (neuro-)symbolic meta-policies effectively steer reinforcement learning agents toward better-aligned and more robust policies, all while offering the crucial benefit of enhanced interpretability.

\paragraph{On the choice of the meta-policy.}
Based on our findings, we can offer concrete recommendations regarding the optimal meta-policy design. The purely neural meta-policy presents the simplest training paradigm as it circumvents the need for a separate symbolic layer and demonstrates strong performance in terms of raw game reward; however, it suffers from limited interpretability and degraded performance when evaluated on simplified games.
In contrast, the purely symbolic meta-policy enhances both interpretability and robustness, though it introduces the overhead of rule definition.
Because this process can be largely automated using LLMs, this minor investment is typically well justified.
Lastly, the \neurosymbolic variant achieves game reward and goal metrics performance comparable to the symbolic meta-policy and is also similarly robust to game simplifications. However, it is more flexible in ambiguous situations.
Based on these results, we recommend the purely symbolic meta-policy for practitioners prioritizing high interpretability and alignment without sacrificing performance. Meanwhile, the \neurosymbolic variation is well-suited for settings with frequent ambiguity.

\paragraph{Scope of interpretability.}
\method provides interpretability at three levels of decreasing transparency: (i)~the meta-policy, whose decisions are fully traceable to a single rule in the symbolic variant and restricted to rule-admissible skills in the neuro-symbolic variant, where the final choice in ambiguous states is made by the learned Q-function; (ii)~the skill objectives, defined by explicit, human-readable reward functions; and (iii)~the low-level action selection within each skill, which remains neural. Our claims therefore concern \emph{why} a skill is selected and \emph{what} it is trained to achieve, not \emph{how} it is executed. Accordingly, our complexity metrics cover only the symbolic layer, and a user study evaluating human understanding of \method agents is left for future work.

\paragraph{Prior knowledge requirements and manual adaptation.}
Two components of \method require prior domain knowledge: the specification of skill reward functions and the definition of meta-policy rules for the symbolic and neurosymbolic variants. While these elements could be engineered manually, we leverage LLMs to minimize the need for human expert intervention. Utilizing the prompts detailed in Appendix~\ref{sec:llm_interaction}, the framework mainly requires the raw game manual.
After obtaining the generated meta-policy and skill rewards, the user is expected to verify correctness of the output, ensure the code respects framework dependencies, and refine thresholds, rule priorities and reward terms where necessary (\cf~\Cref{tab:llm_edits}).

\paragraph{Limitations.}
\method relies on object-centric scene decoders that accurately provide the agent with objects and their positions and sizes from raw images, which we assume to exist in this work.
For Atari games, multiple approaches are viable~\citep{li2017objectsensitive, locatello2020objectcentric, lin2020space, delfosse2023boosting} with some achieving near-perfect sprite extraction \citep{smirnov2021marionette}. 
For real-world data, recent advancements have significantly improved robustness, with models like SAM2~\citep{ravi2025segment} reaching up to 90\% $\mathcal{J}\&\mathcal{F}$ accuracy on zero-shot segmentation.
However, even the best object detector will produce misdetections, so we additionally validate the performance of \method under noisy detections (\cf~Appendix~\ref{apdx:noise}). 
The LLM based generation of reward functions and meta-policies require a pre-defined description of the task, i.e. the game's manual (as done in \cite{wu2023read}).
Moreover, while the generated functions generally capture valid semantics and are logically consistent, the implementations still require framework adaptations and semantic refinements (\cf~\Cref{tab:llm_edits}).
Prior works on LLM-based reward design (e.g., \cite{xie2024t2r}, \cite{ma2024eureka}, \cite{kaufmann2024ocalm}) have documented recurring issues such as reward misspecification, proxy objective selection, and over- or under-constrained preconditions, highlighting the need for careful verification.
NEXUS currently addresses these risks through manual inspection of the generated code.
Extending our framework by incorporating automatic mechanisms, e.g., refinement from trajectory feedback and planning, as explored by SCALAR~\citep{zabounidis2026scalar}, is a crucial direction for enhancing robustness and scalability.
Lastly, the presented approach is based on Q-learning and therefore limited to discrete action spaces, however, extension to continuous action spaces seems viable, e.g., by adopting an off-policy actor critic instead of $\epsilon$-greedy action selection ~\citep{lillicrap2016continuous, haarnoja2018soft}.

%% file: sections/related_work.tex
\section{Related Work}
\label{sec:related_work}

\textbf{Interpretable and Hierarchical Reinforcement Learning.}
Interpretability in RL can be introduced at various stages of the pipeline~\citep{glanois2024survey}, often by deriving symbolic state representations from raw observations via object recognition or segmentation~\citep{li2017objectsensitive, locatello2020objectcentric, kirillov2023segment, lin2020space, delfosse2023boosting}. 
Such object-centric states have enabled interpretable policies through decision trees~\citep{silva2019optimization, likmeta2020combining, delfosse2024interpretable}, logic rules~\citep{maes2012policy, akrour2019towards, delfosse2023interpretable}, parametric functions~\citep{luo2024endtoend} or programmatic policies and trees~\citep{verma2019imitationprojected, anderson2020neurosymbolic, kohler2024interpretable}.
Neural and logical policies 
can also be efficiently combined \citep{shindoblendrl}.
Complementary efforts introduce hierarchical decompositions, where high-level interpretable policies select among low-level sub-policies, leveraging annotated task sketches~\citep{andreas2017modular} or (differentiable) symbolic planning~\citep{leonetti2016synthesis, yang2018peorl, jin2022creativity, lyu2019sdrl, ye2025learning}. 
Hierarchical symbolic planning based on object-centric representations has been shown to be beneficial for task transferability~\citep{james2022autonomous} and robotics applications~\citep{sharma2020learning}.
Our work extends these directions by enforcing semantic separation of skills, introducing a symbolic meta-policy to improve understandability and integrating LLMs throughout the pipeline. 

\paragraph{Reward Shaping and Reward Machines.}
Prior work has explored structured reward specifications to improve learning efficiency. Potential-based reward shaping \citep{ng1999policy} provides a theoretically grounded method for augmenting rewards while preserving optimality, but relies on manually specified potential functions.
Reward machines offer a more expressive automata-based formalism for encoding non-Markovian reward logic \citep{icarte2018using}, with recent work even leveraging LLMs for automated design \cite{castanyer2025armfm}.
However, these approaches typically learn a single value function over a flat action space and have not been evaluated on visually complex domains such as Atari or Crafter.
In contrast, our approach learns a separate Q-function per skill within a hierarchical framework, using object-centric reward specifications to enforce semantic disentanglement. 

\paragraph{LLM-guided skill learning.}
Concurrent to our work, SCALAR~\citep{zabounidis2026scalar} uses LLMs to propose
skills as STRIPS-like operators with preconditions, effects and reward functions, sequences them with a symbolic planner, and trains the corresponding options with PPO in a skill-by-skill curriculum, refining the specifications from rollouts via trajectory analysis. While SCALAR targets sample-efficient learning of long
prerequisite chains in crafting domains and executes skills in a fixed planned order, NEXUS targets interpretable high-level decision making: its (neuro-)symbolic meta-policy re-selects skills at every step from the object-centric state, all skills are learned concurrently from shared off-policy experience without state resets, and skill rewards are kept separate from the environment reward to enforce disentanglement. The two directions are complementary; SCALAR's trajectory-based refinement could automate the verification of NEXUS's LLM-generated rewards and meta-policies.
\paragraph{Relation to the Options framework.}
HRL has been studied extensively, with the Options framework \citep{sutton1999between} as the most prominent formulation.  
\method\ can be interpreted as an instantiation of the Options framework with intra-option learning, where multiple options are learned simultaneously from shared off-policy experience, using sub-policies guided by option-specific reward functions and coordinated by a neuro-symbolic meta-policy.
Key differences arise in temporal abstraction and in the treatment of initiation and termination conditions.  
Rather than executing an option until termination, the meta-policy selects the active option at every time step, jointly determining activation and termination.  
The logical rule set as the meta-policy serves as a generalized initiation set, enabled by the object-centric encoding of observations.

%% file: sections/conclusion.tex
\section{Conclusion}
In this work, we present \method, a hierarchical RL method that combines high interpretability on meta-policy level with neural, low-level action execution.
The evaluation suggest several advantages of our approach. 
It learns disentangled sub-policies corresponding to identifiable skills, provides high-level interpretable and modular structures for inspection and intervention, reduces reward-hacking through integration of domain priors, and remains robust to environment simplifications where standard deep RL agents fail. 
We also found that LLMs can successfully be integrated into the \method pipeline to generate skill decompositions, reward functions, and symbolic meta-policies, enabling interpretability and dynamic adaptation to simplified environments. 
Future work should increase the autonomy of policy adaptation by incorporating mechanisms to detect when new skills or meta-policy updates are required, \eg via causal world models~\citep{yang2025towards, dillies2025better}, which could enable scaling to complex, open-ended environments. Updates to the skills or meta-policy could be retrieved by re-querying an LLM.
In conclusion, this work advances RL interpretability and modularity through symbolic and object-centric reasoning while supporting human-in-the-loop control at the skill level, offering a promising path toward transparent, adaptable, and aligned agents.

\paragraph{Broader Impact.}
This work advances RL interpretability and modularity by integrating symbolic structure and object-centric reasoning. 
Beyond enhanced transparency, it enables human-in-the-loop control: practitioners may inspect, modify, or constrain agent behavior at the skill level.
We believe that this precise, skill-level alignment and the flexible adaptation are among the most promising avenues opened by this approach.
While LLM-generated rewards and meta-policies are viable, they remain susceptible to misspecification and subsequent undesirable actions, making human oversight essential for safety-critical applications.

%% file: sections/appendix.tex
\section{Motivation - Shortcomings of existing approaches}
\label{apdx:motiv}

\begin{figure}[H]
    \centering
    \includegraphics{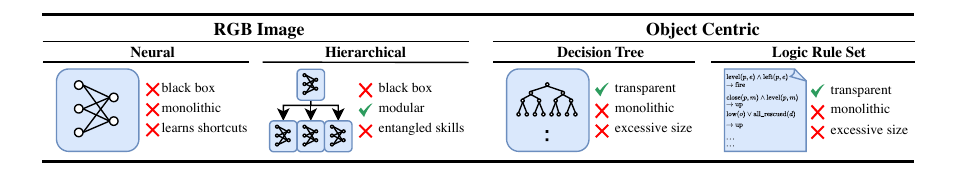}
    \caption{\textbf{Existing approaches are not interpretable.} Neural RL agents are black-boxes and learn shortcuts. Hierarchical RL remains a black box and typically produces entangled skills. Symbolic agents are transparent, but grow massively with policy complexity.}
    \label{fig:enter-label}
\end{figure}

\subsection{Interpretable Policy Sizes}
We visualize the actual logic policies retrieved from existing interpretable RL methods in \Cref{lst:scobots} and \Cref{lst:nudge}. While these policies have maximum transparency on low-level action selection, they are difficult to interpret for humans due to the massive size.
\method on the other hand presents a simple solution by incorporating hierarchical abstraction and thus allowing for compact and interpretable high-level policies, \cf~\Cref{lst:symbolic_meta_pol} and \Cref{lst:nesy_meta_pol}.

\input{sections/logic_policies}

\section{Hierarchical PQN Algorithm}
\label{apdx:hpqn}
The complete algorithm for hierarchical PQN with a neural meta-policy is provided in \Cref{alg:h_pqn_full}.

\begin{algorithm*}
\caption{Hierarchical PQN}
\begin{algorithmic}[1]
\REQUIRE Update period $U$, number of parallel environments $E$, number of skills $N$, exploration probability $\epsilon$
\ENSURE Learned Q-network parameters $\{\phi_n\}_{n=1}^N$, $\phi_{\text{meta}}$
\STATE Initialize Q-network parameters $\{\phi_n\}_{n=1}^N$, $\phi_{\text{meta}}$
\STATE Sample initial states $s_0^e \sim P_0$ for $e \in \{0, \dots, E-1\}$
\STATE $t \gets 0$
\FOR{each episode}
    \FORALL{$e \in \{0, \dots, E-1\}$ \textbf{in parallel}}
        \STATE Sample skill $\pi_t^e \sim \pi_{\text{meta}}$
        \STATE With probability $\epsilon$: $a_t^e \sim \text{Unif}$, else $a_t^e \sim \pi_t^e$
        \STATE Sample rewards $r_t^{e} \sim P_{R}(s_t^e, a_t^e)$, skill rewards $r_t^{e,n} \sim P_{R,n}(s_t^e, a_t^e)$ for all $n$
        \STATE Sample next state $s_{t+1}^e \sim P_S(s_t^e, a_t^e)$
        \STATE $t \gets t + 1$
    \ENDFOR
    \IF{$t \bmod U = 0$}
        \STATE Compute meta $\lambda$-returns $R_{\lambda, t-1}^e$ to $R_{\lambda, t-U}^e$ for all $e$
        \STATE Compute skill $\lambda$-returns $R_{\lambda, t-1}^{e,n}$ to $R_{\lambda, t-U}^{e,n}$ for all $e$, $n$
        \FOR{number of epochs}
            \FOR{number of minibatches}
                \STATE Sample minibatch $B$ of size $b \leq EU$ from $\{(t-U, 0), \dots, (t-1, E-1)\}$
                \STATE Update meta:
                \[
                \phi_{\text{meta}} \gets \phi_{\text{meta}} + \frac{\alpha_t}{2b} \nabla_{\phi_{\text{meta}}} \sum_{(j,\tau) \in B} \left(R_{\lambda, \tau}^j - Q_{\phi_{\text{meta}}}(s_\tau^j)\right)^2
                \]
                \STATE Update skills:
                \[
                \phi_n \gets \phi_n + \frac{\alpha_t}{2b} \nabla_{\phi_n} \sum_{(j,\tau) \in B} \left(R_{\lambda, \tau}^{j,n} - Q_{\phi_n}(s_\tau^{j,n})\right)^2, \quad \forall n \in \{1, \dots, N\}
                \]
            \ENDFOR
        \ENDFOR
    \ENDIF
\ENDFOR
\end{algorithmic}
\label{alg:h_pqn_full}
\end{algorithm*}

\section{Implementation Details}
\label{apdx:Impl}

\paragraph{Setup.}
We base our implementations on CleanRL~\citep{huang2022cleanrl} and PureJaxRL~\citep{lu2022discovered}, adapting them to object-centric inputs by replacing convolutional encoders with lightweight MLPs for feature extraction. Hyperparameters are listed in \Cref{table:hps_1} and \Cref{table:hps_2} and remain largely consistent with the original implementations, except for an increased number of parallel environments enabled by the efficiency of JAX-based code.
\\
The exploration parameter \( \epsilon \) was selected via a brief hyperparameter sweep in the range \([1, 0.001]\), using final test return as the selection criterion.
\\
Each experiment is run with six random seeds (0-5) to ensure robust reporting. Reported plots include the corresponding standard deviation.
\\
All experiments were conducted on a single NVIDIA Tesla V100-SXM3-32GB-H GPU on an NVIDIA DGX Server (Version 5.1.0) with CUDA 12.4.
\\ Our code repository\footnote{\url{https://github.com/remunds/symbolic_options}} includes all necessary code and config files to reproduce the results from the experiments section, including plots.

\subsection{Environments}
\label{apdx:envs}
\begin{figure}[t]
    \centering
    \begin{subfigure}[b]{0.32\textwidth}
        \includegraphics[width=\textwidth]{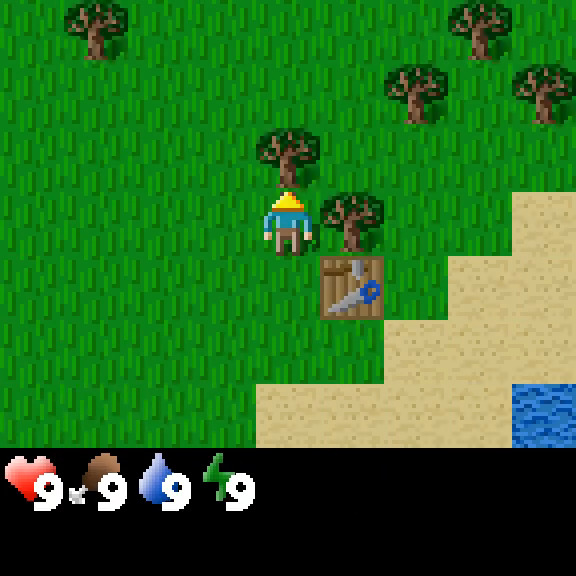}
        \caption{Crafter}
    \end{subfigure}
    \hfill
    \begin{subfigure}[b]{0.32\textwidth}
        \includegraphics[width=\textwidth]{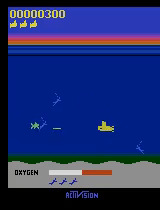}
        \caption{Seaquest}
    \end{subfigure}
    \hfill
    \begin{subfigure}[b]{0.32\textwidth}
        \includegraphics[width=\textwidth]{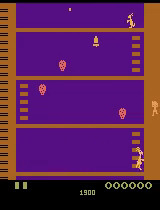}
        \caption{Kangaroo}
    \end{subfigure}
    \caption{The evaluation environments: Crafter, Seaquest and Kangaroo.}
    \label{fig:envs}
\end{figure}

Screenshots of the environments Crafter, Seaquest and Kangaroo are available in \Cref{fig:envs}.

These games are selected because their gameplay naturally decomposes into low-level skills that must be combined to solve the overall task.
In Kangaroo, the objective is to ascend through platforms while avoiding enemies and collecting berries.
In Seaquest, the player navigates a submarine to rescue divers and return them to the surface, while avoiding hazards such as sharks and enemy submarines.
These two games are known to exhibit \textit{reward hacking} behavior, meaning that the agents find simple shortcuts of increasing their return rather than the intended way that humans would typically choose.
Examples include exclusively shooting enemies in Seaquest and only catching falling apples or boxing monkeys in Kangaroo.

Unlike Atari games, Crafter presents a more complex, open-ended challenge that requires agents to acquire and coordinate multiple skills, including navigation, combat, resource collection, and crafting.

\subsection{Evaluation Metrics}
We empirically evaluate agent performance using three metrics: (1) \textbf{Skill Returns} to test whether the skills were learned successfully, (2) \textbf{Human-Normalized Score} (HNS) for absolute performance relative to human and random baselines and (3) \textbf{Aligned Environment Goal Scores} that measure performance based on the main goals described in the game's manual.

\paragraph{Human Normalized Score.}
Human-Normalized Score standardizes agent performance across Atari environments by accounting for differences in reward scales~\citep{mnih2015humanlevel, MachadoBTVHB18}.
Given the average agent score \( A \), human score \( H \), and random score \( R \), HNS is defined as:
\[
\text{HNS} = \frac{A - R}{|H - R|}
\]
A value of 1.0 indicates human-level performance, values greater than 1.0 indicate superhuman performance, and values below 0 denote sub-random behavior.
We adopt the human and random baselines from~\citet{badia2020agent}, derived from professional human play.

\paragraph{Aligned Environment Goal Scores.}

Environment reward signals may not always align with the intended task objectives and can be susceptible to reward hacking.
In such cases, agents may learn high-reward behaviors that are non-intuitive and deviate from human-like solutions.
To better capture progress toward the actual environment goals, we define two aligned goal-based metrics grounded in the objectives stated in the game manuals. 

For \textit{Seaquest}, the goal is to rescue as many divers as possible; for \textit{Kangaroo}, it is to help the mother kangaroo reach and rescue her baby, located on the topmost platform. Accordingly, we track the number of divers retrieved and the number of platforms reached, respectively.

\begin{table}[tb]
\centering
\caption{Atari Hyperparameters}
\label{table:hps_1}
\begin{tabular}{lll}
\toprule
\textbf{Parameter} & \textbf{PQN(-based)} & \textbf{PPO} \\
\midrule
Total Timesteps & $5 \times 10^7$ & $5 \times 10^7$ \\
Num Environments & 1024 & 128 \\
Num Steps per Update & 32 & 32 \\
Learning Rate & $2.5 \times 10^{-4}$ & $2.5 \times 10^{-4}$ \\
Max Grad Norm & 10 & 0.5 \\
Discount Factor ($\gamma$) & 0.99 & 0.99 \\
GAE Lambda ($\lambda$) & 0.65 (0.5 with learned meta) & 0.95 \\
GAE Meta Lambda ($\lambda$) & 0.9 & -- \\
Num Epochs & 1 & 2 \\
Num Minibatches & 256 & 4 \\
Hidden Size & 64 & -- \\
Num Layers & 3 & -- \\
Normalization & Layer Norm & -- \\
Clip $\epsilon$ & -- & 0.2 \\
Entropy Coef & -- & 0.01 \\
Value Function Coef & -- & 0.5 \\
Anneal LR & False & True \\
$\epsilon$-Start/End/Decay & 1.0 / 0.001 / 0.1 & -- \\
Meta $\epsilon$-Start/End/Decay & 1.0 / 0.001 / 0.3 & -- \\
\bottomrule
\end{tabular}
\end{table}

\begin{table}[tb]
\centering
\caption{Crafter Hyperparameters}
\label{table:hps_2}
\begin{tabular}{lll}
\toprule
\textbf{Parameter} & \textbf{PQN(-based)}\\
\midrule
Total Timesteps & $1 \times 10^9$ \\
Num Environments & 512\\
Num Steps per Update & 128\\
Learning Rate & $3.0 \times 10^{-4}$\\
Max Grad Norm & 0.5\\
Discount Factor ($\gamma$) & 0.99 \\
GAE Lambda ($\lambda$) & 0.5 \\
GAE Meta Lambda ($\lambda$) & 0.5 \\
Num Epochs & 4\\
Num Minibatches & 4 \\
Hidden Size & 512 \\
Num Layers & 1 \\
Normalization & Layer Norm \\
Anneal LR & True \\
$\epsilon$-Start/End/Decay & 1.0 / 0.005 / 0.1 \\
Meta $\epsilon$-Start/End/Decay & 1.0 / 0.005 / 0.1 \\
\bottomrule
\end{tabular}
\end{table}

\subsection{Interpretability Proxies - Detailed Computation}
\label{apdx:interpret_measures}

To rigorously compare policy complexity between NUDGE~\citep{delfosse2023interpretable} and \method, we adopt the formal rule-based complexity metrics defined by \cite{lakkaraju2016interpretable} alongside Average Decision Path Length \citep{wu2017beyond}:
\begin{itemize}
    \item \textbf{Rule Size} ($|\mathcal{R}|$): The total number of distinct rules or conditional logic branches comprising the policy.
    \item \textbf{Total Rule Length} ($\mathcal{L}_{\text{total}}$): The sum of all body literals (conditions) evaluated across the entire policy ($\mathcal{L}_{\text{total}} = \sum_{r \in \mathcal{R}} |r|$).
    \item \textbf{Average Decision Path Length} ($\text{ADPL}$): The expected number of condition checks executed to make a single decision under a uniform distribution over execution paths \cite{wu2017beyond}.
\end{itemize}

---

\subsubsection*{1. Metric Definitions}
\begin{align*}
|\mathcal{R}| &= \text{Total count of rules/branches in policy } \mathcal{R} \\
\mathcal{L}_{\text{total}} &= \sum_{r \in \mathcal{R}} |r| \quad \text{where $|r|$ is the number of body literals in rule $r$} \\
\text{ADPL} &= \frac{1}{|\mathcal{K}|} \sum_{k=1}^{|\mathcal{K}|} C_k \quad \text{where $C_k$ is the cumulative evaluations to trigger path $k$}
\end{align*}

---

\subsubsection*{2. NUDGE Policy Computations}
NUDGE for Seaquest (\cf \Cref{lst:nudge}) consists of $|\mathcal{R}| = 16$ sequential Prolog rules evaluated top-to-bottom.

\subsubsection*{A. Rule Size and Total Rule Length}
\begin{itemize}
    \item \textbf{Rule Size ($|\mathcal{R}|_{\text{NUDGE}}$):} $\mathbf{16 \text{ rules}}$
    \item \textbf{Total Rule Length ($\mathcal{L}_{\text{NUDGE}}$):}
    \begin{align*}
    \mathcal{L}_{\text{NUDGE}} &= \sum_{i=1}^{16} |r_i| \\
    &= 1 + 1 + 4 + 4 + 4 + 4 + 2 + 2 + 3 + 3 + 3 + 3 + 3 + 3 + 3 + 3 \\
    &= \mathbf{46 \text{ literals}} 
    \end{align*}
\end{itemize}

\subsubsection*{B. Average Decision Path Length ($\text{ADPL}_{\text{NUDGE}}$)}
Evaluating sequentially, reaching rule $k$ requires evaluating and failing all conditions in rules $1 \dots k-1$ plus evaluating rule $k$. The cumulative cost $C_k = \sum_{i=1}^k |r_i|$ for each rule $k \in \{1, \dots, 16\}$ is:
\begin{align*}
C_1 &= 1, \quad C_2 = 2, \quad C_3 = 6, \quad C_4 = 10, \quad C_5 = 14, \quad C_6 = 18, \quad C_7 = 20, \quad C_8 = 22, \\
C_9 &= 25, \quad C_{10} = 28, \quad C_{11} = 31, \quad C_{12} = 34, \quad C_{13} = 37, \quad C_{14} = 40, \quad C_{15} = 43, \quad C_{16} = 46.
\end{align*}

Summing cumulative checks across all 16 outcomes:
\begin{equation*}
\sum_{k=1}^{16} C_k = 377 \implies \text{ADPL}_{\text{NUDGE}} = \frac{377}{16} = \mathbf{23.56 \text{ evaluations}}
\end{equation*}

---

\subsubsection*{3. \method (symbolic) Computations}
\Cref{lst:symbolic_meta_pol} defines an explicit control-flow hierarchy over 4 execution branches.

\subsubsection*{A. Rule Size and Total Rule Length}
\begin{itemize}
    \item \textbf{Rule Size ($|\mathcal{R}|_{\text{NEXUS-1}}$):} $\mathbf{4 \text{ execution branches}}$ (3 conditional statements + 1 default fallback)
    \item \textbf{Total Rule Length ($\mathcal{L}_{\text{NEXUS-1}}$):}
    \begin{align*}
    \mathcal{L}_{\text{NEXUS-1}} &= |r_1| + |r_2| + |r_3| + |r_4| \\
    &= 2 + 1 + 1 + 0 \\
    &= \mathbf{4 \text{ conditions}}
    \end{align*}
\end{itemize}

\subsubsection*{B. Average Path Length ($\text{ADPL}_{\text{NEXUS-1}}$)}
Accounting for short-circuit evaluation across the 4 possible decision paths ($\mathcal{K} = \{1, 2, 3, 4\}$), where a satisfied disjunction requires one check and a failed disjunction two:
\begin{itemize}
    \item Path 1 (\texttt{surface}): $C_1 = 1$ check
    \item Path 2 (\texttt{fight\_enemies}): $C_2 = 2 + 1 = 3$ checks
    \item Path 3 (\texttt{rescue\_divers}): $C_3 = 2 + 1 + 1 = 4$ checks
    \item Path 4 (\texttt{fight\_enemies} default): $C_4 = 2 + 1 + 1 = 4$ checks
\end{itemize}

\begin{equation*}
\text{ADPL}_{\text{NEXUS-1}} = \frac{1 + 3 + 4 + 4}{4} = \mathbf{3.00 \text{ evaluations}}
\end{equation*}

---

\subsubsection*{4. \method (nesy) Computations}
\Cref{lst:nesy_meta_pol} evaluates two independent conditional blocks sequentially; \texttt{fight\_enemies} is always admissible.

\subsubsection*{A. Rule Size and Total Rule Length}
\begin{itemize}
    \item \textbf{Rule Size ($|\mathcal{R}|_{\text{NEXUS-2}}$):} $\mathbf{2 \text{ rule blocks}}$
    \item \textbf{Total Rule Length ($\mathcal{L}_{\text{NEXUS-2}}$):}
    \begin{align*}
    \mathcal{L}_{\text{NEXUS-2}} &= |r_1| + |r_2| \\
    &= 1 + 2 \\
    &= \mathbf{3 \text{ conditions}}
    \end{align*}
\end{itemize}

\subsubsection*{B. Average Decision Path Length ($\text{ADPL}_{\text{NEXUS-2}}$)}
Both blocks execute on every forward pass:
\begin{itemize}
    \item \textbf{Block 1} (\texttt{if diver\_available}): Always $1.0$ check.
    \item \textbf{Block 2} (\texttt{if oxygen\_low or all\_collected}): Evaluates \texttt{oxygen\_low} ($1$ check) + $50\%$ chance of evaluating \texttt{all\_collected} ($+0.5$ check) $= 1.5$ checks.
\end{itemize}

\begin{equation*}
\text{ADPL}_{\text{NEXUS-2}} = 1.0 + 1.5 = \mathbf{2.50 \text{ evaluations}}
\end{equation*}

\clearpage

\section{Additional Results}
\label{apdx:add_results}

\subsection{Comparative evaluation against reward shaping baselines}
\label{apdx:shaped_rewards}

To isolate the benefits of NEXUS, we evaluate its performance against PQN trained on shaped rewards.
To ensure a rigorous comparison, we designed three distinct reward-shaping strategies:

\begin{enumerate}
\item \textbf{PQN + shaped (symbolic):} Utilizes the same LLM-designed symbolic meta-policy as NEXUS (symbolic) to dynamically select the active skill reward at each time step, functioning analogously to a reward machine \citep{icarte2018using, castanyer2025armfm}.
\item \textbf{PQN + shaped (uniform):} Applies a static, equal weighting across all three skill rewards throughout training.
\item \textbf{PQN + shaped (LLM-weighted):} Leverages LLM-generated scalar weights to combine the individual skill rewards into a single fixed objective.
\end{enumerate}

Across both environments, all three reward-shaping baselines fall significantly short of NEXUS in terms of overall performance.
Although these baselines successfully acquire specific localized behaviors, such as \texttt{rescue\_divers} or \texttt{surface}, in \textit{Seaquest}, they focus on single skills and fail to effectively coordinate them, resulting in low overall environment returns.
These results demonstrate that simply augmenting RL methods with shaped or LLM-designed rewards is insufficient, whereas NEXUS's neuro-symbolic architecture enables equal skill learning and coordination.

\begin{figure}[tb]
    \centering
    \includegraphics[width=\linewidth]{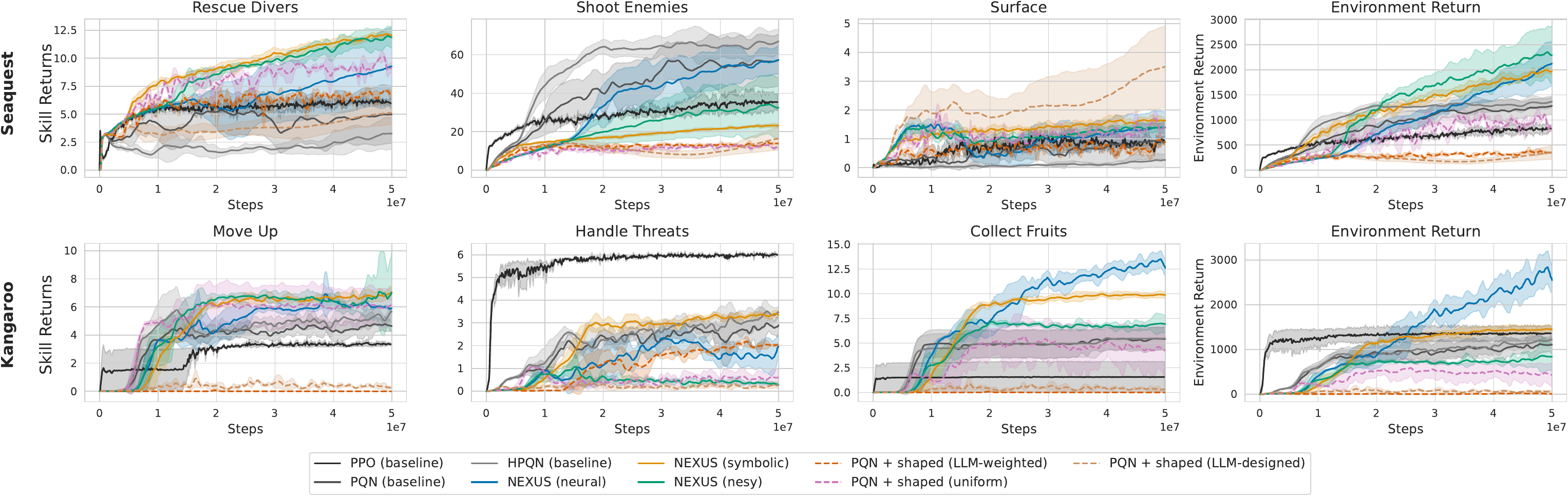}
    \caption{\textbf{Reward-shaping is not sufficient.} Replacing NEXUS's neuro-symbolic meta-policy with LLM-designed or statically weighted reward shaping leads to severe performance degradation across both environments. While baselines successfully acquire isolated skills (e.g., \texttt{surface} in Seaquest), they fail to learn all skills equally well or to coordinate them---demonstrating that NEXUS's skill disentanglement is essential for achieving high overall performance.}
    \label{fig:shaped_rewards}
\end{figure}

\subsection{Analysis of neuro-symbolic meta-policy impact and skill activation dynamics}
\label{apdx:nesy_analysis}

\begin{figure}[tb]
    \centering
    \includegraphics[width=\linewidth]{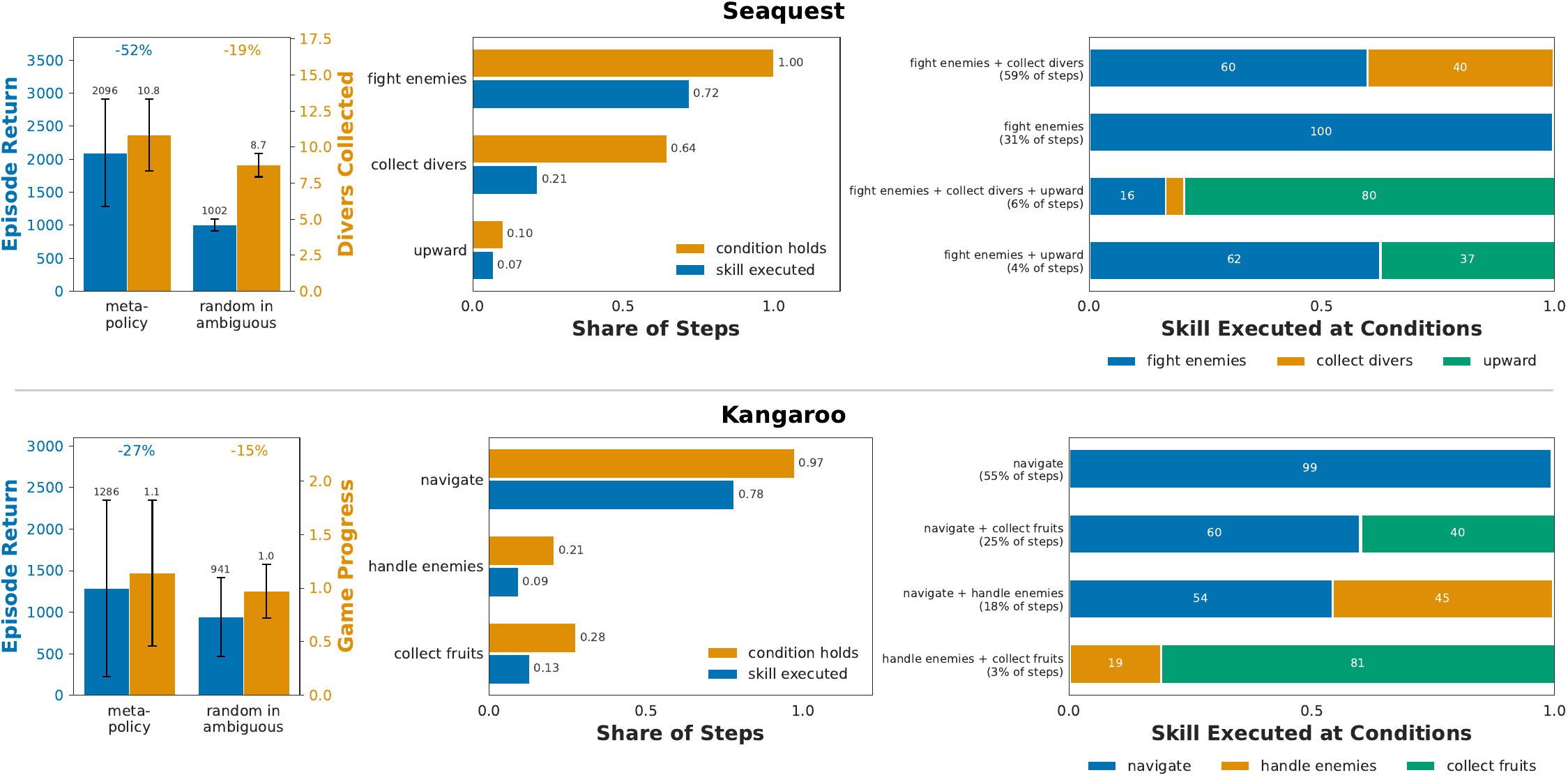}
    \caption{\textbf{Skill activation dynamics and performance impact of the neuro-symbolic meta-policy in \textit{Seaquest} and \textit{Kangaroo}.} Replacing the meta-policy with random skill choices significantly degrades performance, demonstrating its critical role in maximizing returns. Default skills (\texttt{fight\_enemies} in \textit{Seaquest}, \texttt{navigate} in \textit{Kangaroo}) dominate usage, while specialized skills trigger selectively based on context: \texttt{collect\_divers} (20\% total / 30\% eligible) and \texttt{upward} (7\% total / 70\% eligible) in \textit{Seaquest}, versus \texttt{collect\_fruits} (13\% total / 46\% eligible) and \texttt{handle\_enemies} (9\% total / 40\% eligible) in \textit{Kangaroo}.}
    \label{fig:activations_nesy}
\end{figure}

To evaluate the contribution of the neuro-symbolic meta-policy, we analyze both the performance impact of its decision-making mechanism and the activation frequencies of individual skills across Seaquest and Kangaroo (\Cref{fig:activations_nesy}).
We first test the necessity of the Meta-Q-value maximization by replacing it in ambiguous situations where multiple skill conditions evaluate to true with a uniformly random choice over the valid skills.
To prevent frequent switching, random skill selections are held for a minimum duration of 20 steps.
We observe a notable drop in overall environment return and aligned evaluation metrics in both domains, with Seaquest exhibiting a stronger performance degradation than Kangaroo.
This drop confirms that the neuro-symbolic meta-policy plays a critical role in selecting context-appropriate actions rather than relying on arbitrary execution.
Analyzing skill activation frequencies reveals that gameplay in both environments is heavily anchored by dedicated default skills: \texttt{fight\_enemies} in Seaquest and \texttt{navigate} in Kangaroo.
The meta-policy selectively departs from these defaults only when specific conditions permit alternative skills to maximize expected returns.
In Seaquest, the agent heavily balances default combat with targeted objectives.
The \texttt{collect\_divers} skill activates in roughly 20\% of all execution steps, corresponding to 30\% of the steps where it was eligible.
Conversely, the \texttt{upward} skill accounts for only 7\% of total gameplay steps, but triggers in 70\% of all valid opportunities, indicating highly efficient, situational utilization.
In Kangaroo, the \texttt{navigate} skill overwhelmingly dominates total runtime.
Secondary skills execute far more conservatively: \texttt{collect\_fruits} accounts for 13\% of all steps (46\% of eligible opportunities), while \texttt{handle\_enemies} is invoked in just 9\% of total steps (40\% of eligible opportunities).

\subsection{Additional environments}
\begin{figure*}[tb]
    \centering
    \includegraphics[width=\linewidth]{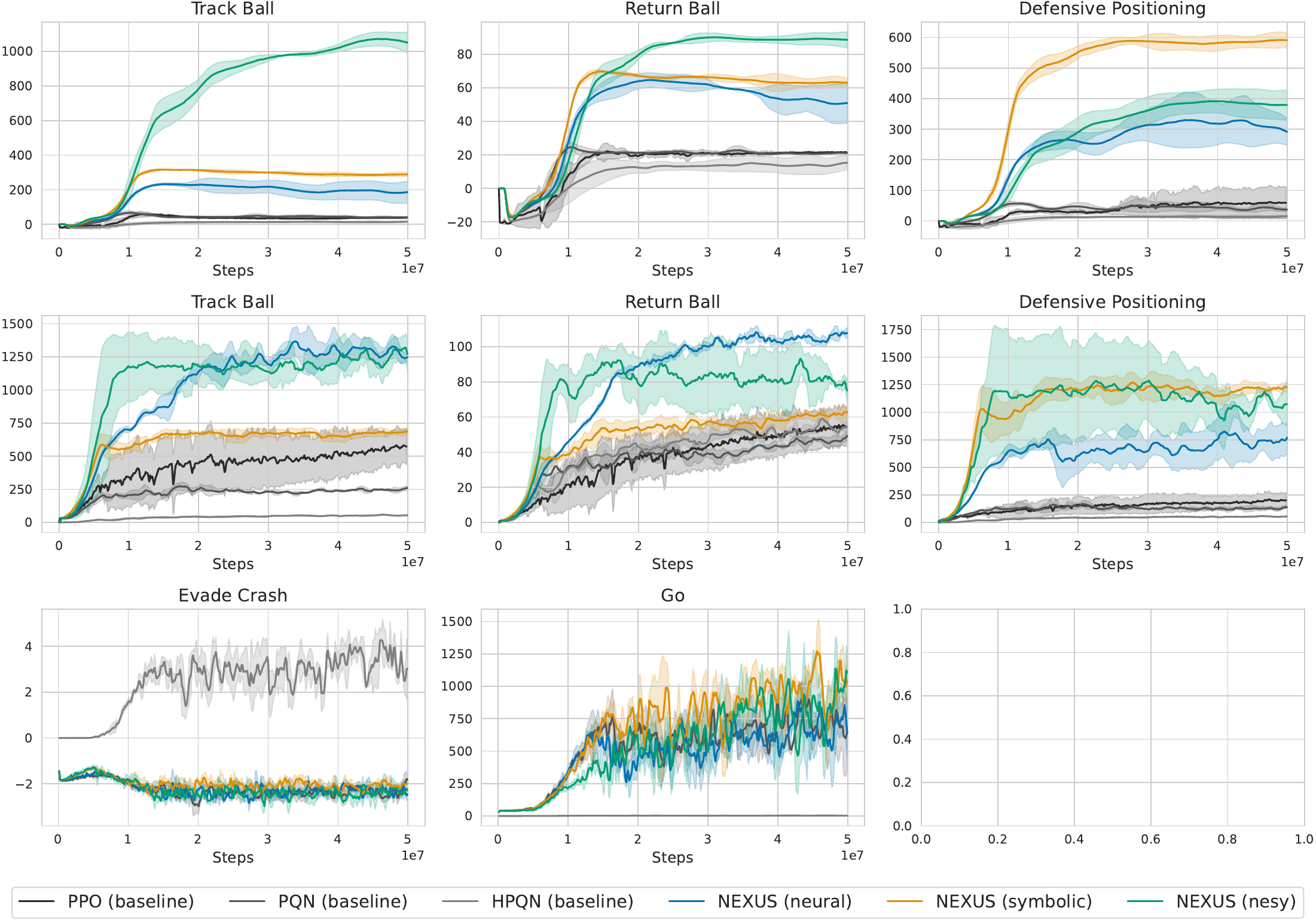}
    \caption{\method successfully learns skills in \textit{Pong} (top), \textit{Breakout} (middle), and \textit{Freeway} (bottom).}
    \label{fig:skill_learn_rest}
\end{figure*}

\begin{figure*}[tb]
    \centering
    \includegraphics[width=\linewidth]{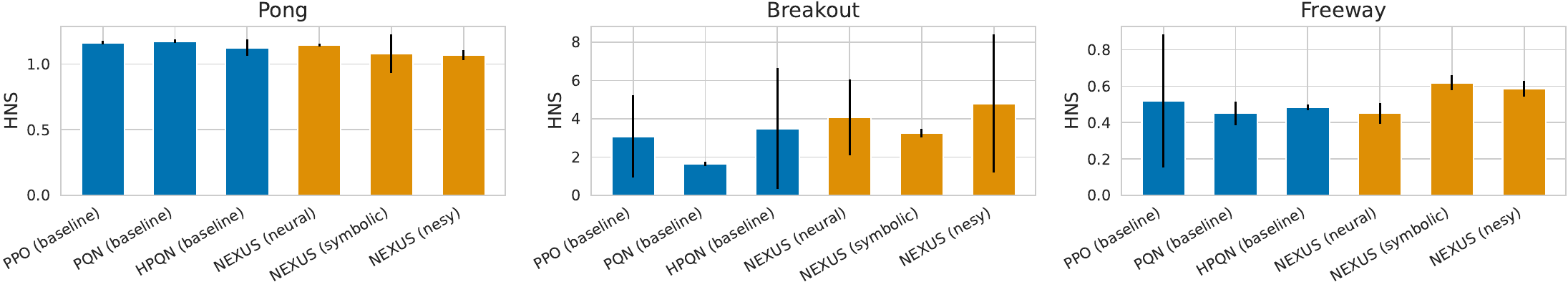}
    \caption{\method achieves performance comparable to baseline methods across the three evaluated environments.}
    \label{fig:single_metric}
\end{figure*}

To assess the generality of \method, we extend our evaluation to three additional Atari environments: \textit{Pong}, \textit{Breakout}, and \textit{Freeway}.
Skill learning curves for these environments are shown in \Cref{fig:skill_learn_rest}, while comparisons to baseline agents and ablations—evaluated via human-normalized scores—are presented in \Cref{fig:single_metric}.
Note that these games are generally less complex than \textit{Kangaroo} and \textit{Seaquest}, and the learned skills are not strictly necessary to achieve the environment goals. 
In particular, the skills "Move Up" and "Avoid Crash" in \textit{Freeway} largely correspond to atomic actions such as \texttt{forward} or \texttt{noop}. 
As such, a simple fixed meta-policy operating directly on primitive actions could suffice for solving this task.

\subsection{Neuro-symbolic RL baselines.}
\label{apdx:nesy_baselines}
\begin{figure}[tb]
    \centering
    \includegraphics[width=0.8\textwidth]{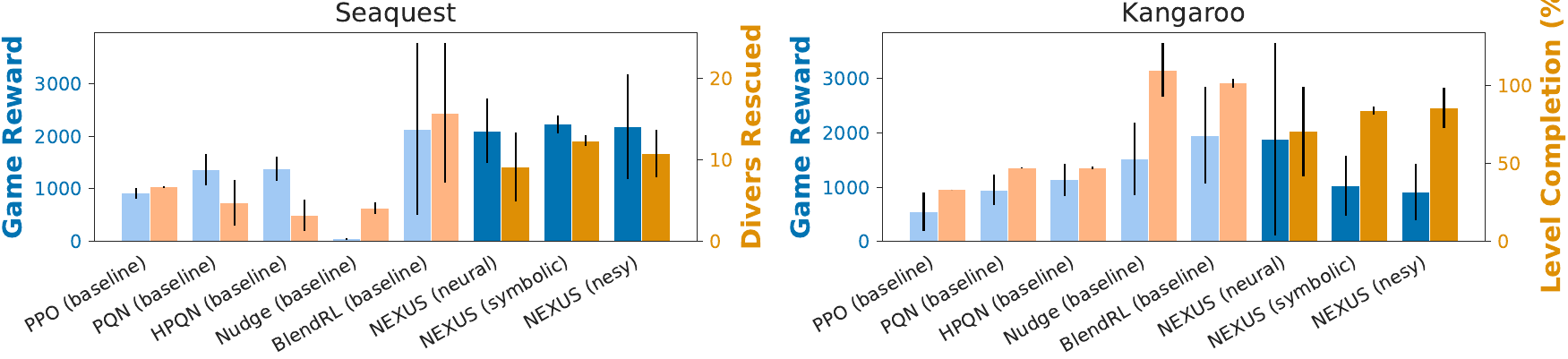}
    \caption{Comparison to other neuro-symbolic RL methods.}
    \label{fig:baselines_1}
\end{figure}

\begin{figure}[tb]
    \centering
    \includegraphics[width=0.8\textwidth]{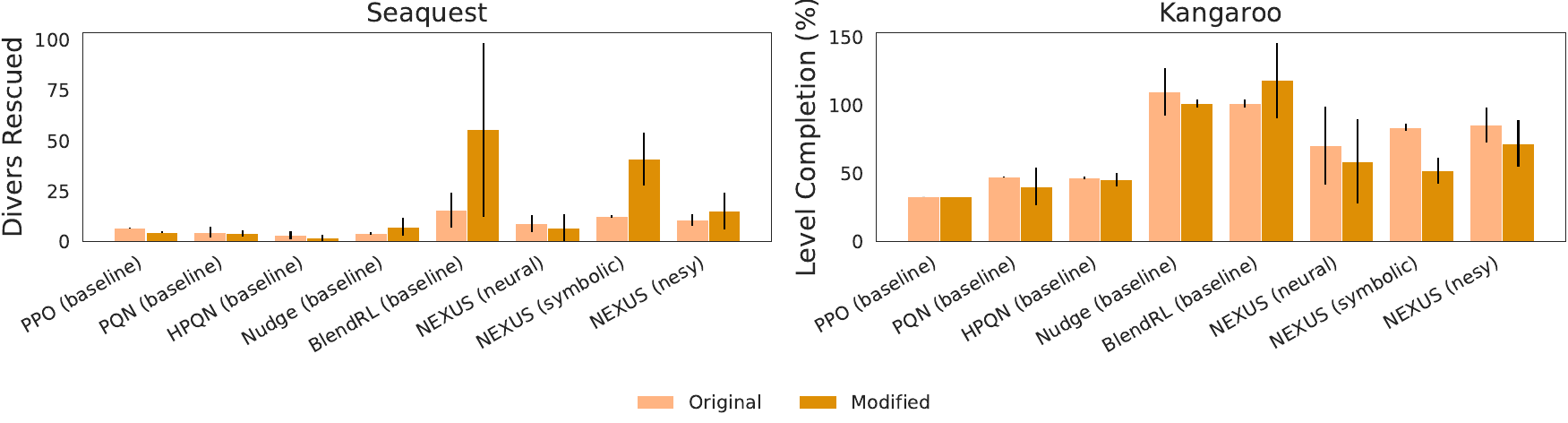}
    \caption{Comparison to other neuro-symbolic RL methods on simplified games.}
    \label{fig:baselines_2}
\end{figure}

For improved comparison, we provide baseline scores of existing interpretable and \neurosymbolic methods on both the default games (\cf~\Cref{fig:baselines_1}) and the simplifications (\cf~\Cref{fig:baselines_2}). We compare to NUDGE \citep{delfosse2023interpretable} and BlendRL \citep{shindoblendrl}. 
All scores, together with the reported scores from SCoBots \citep{delfosse2024interpretable} (game simplification scores from \cite{delfosse2025deep}) are reported in \Cref{tab:performance_comp}.
While prior neuro-symbolic methods achieve competitive scores, they suffer from significant limitations regarding transparency and flexibility. 
Primarily, interpretability in these baselines is restricted to low-level actions. Consequently, the overall behavioral policy must be defined at the action granularity, resulting in excessive complexity that is difficult to parse (\cf~\Cref{apdx:motiv}). Furthermore, these approaches lack explicit representations of the high-level objectives currently being pursued (e.g., `collect all divers').

The logical rules within NUDGE \citep{delfosse2023interpretable} and BlendRL \citep{shindoblendrl} are static and strictly followed, resulting in rigid policies. While BlendRL attempts to mitigate this via a switching mechanism that activates a neural policy, this compromise diminishes transparency; the neural component maximizes extrinsic reward rather than adhering to an interpretable sub-goal.

Moreover, all baseline approaches \textbf{necessitate substantial manual engineering}. NUDGE and BlendRL require the definition of specific valuation functions based on detected objects and attributes, while SCoBots \citep{delfosse2024interpretable} relies on the manual definition and selection of relational functions. 
In contrast, \method operates at the abstracted level of skills, facilitating the interpretation and manipulation of the agent's active goals. As a result, the meta-policy definitions in \method are significantly more concise than the action-level policies of SCoBots, NUDGE, and BlendRL. Additionally, \method eliminates the need for manual valuation functions by autonomously learning skills via LLM-generated reward functions.

Finally, the performance of NUDGE and BlendRL in simplified environments is largely attributable to extensive hard-coded logic and specific valuation functions (e.g. `fire if same\_depth(player, enemy) and visible(enemy) and facing\_right(player) and left\_of(player, enemy)') . Deriving these rigid rules is significantly more labor-intensive than generating meta-policy and reward functions in \method, rendering the baselines less scalable to complex environments. This suggests an inherent trade-off between high-level interpretability, manual labor, and performance. While previous algorithms prioritize maximum action-level transparency and performance by incorporating extensive manual inductive biases, this work targets high-level interpretability with a significantly reduced reliance on such biases.

\input{sections/baseline_comp_table}

\section{Noisy Detections}
\label{apdx:noise}

While object detection methods become increasingly reliable, misdetections still happen.
To simulate unreliable object extractors, we incorporate a misdetection chance and gaussian noise to each object attribute during both training and testing.
We conduct experiments with 10\% misdetection rate and 3px gaussian noise. 
To emulate the effect of a kalman filter estimating the objects movements during each step, instead of zeroing out the detected objects, we keep the previous time step detections.
On top of that, we add gaussian noise with a standard deviation of 3 at each attribute.

\begin{figure}[tb]
    \centering
    \includegraphics[width=0.8\textwidth]{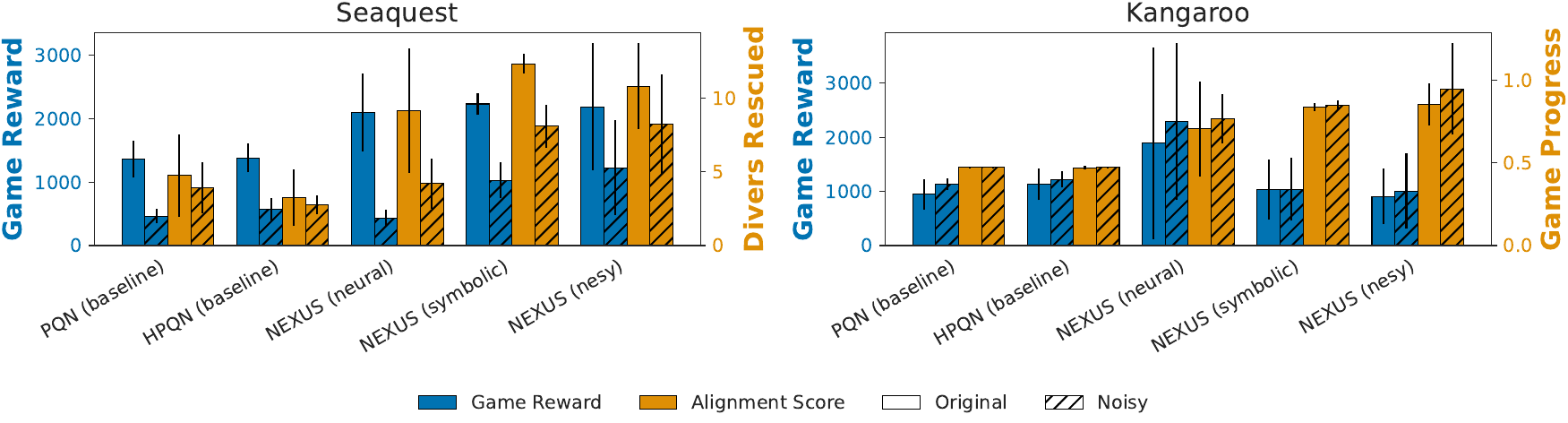}
    \caption{Noisy evaluation under 10\% misdetection and noise.}
    \label{fig:noisy_evaluation}
\end{figure}

We observe significant differences between Seaquest and Kangaroo.
For Seaquest, when introducing noise only during evaluation (\Cref{fig:noisy_evaluation}), all methods lose performance significantly. However, both the symbolic and 
\neurosymbolic approaches keep rescuing a relatively high number of divers.
In Kangaroo, on the other hand, 
the experiments surprisingly indicate that all approaches increase in performance.

\begin{figure}[tb]
    \centering
    \includegraphics[width=0.8\textwidth]{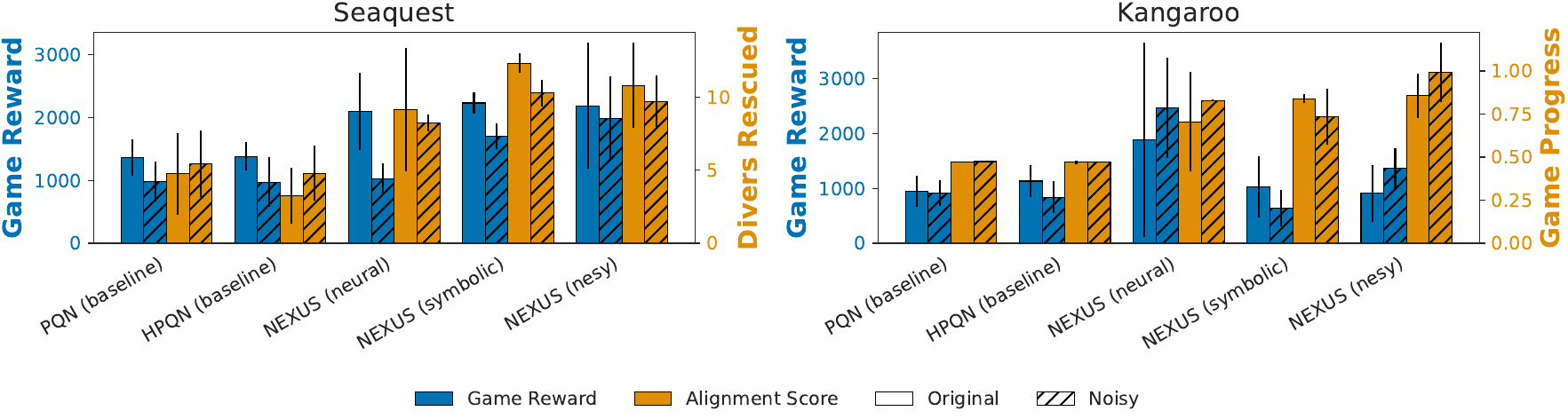}
    \caption{Noisy training and evaluation under 10\% misdetection and noise.}
    \label{fig:noisy_train_and_evaluation}
\end{figure}

The performance drops show different characteristics when the approaches also experience noise during training (\Cref{fig:noisy_train_and_evaluation}).

In Seaquest, all NEXUS approaches drop only a small amount and remain reliable. 
We again observe only small performance drops (and even improvements) in Kangaroo.
Our working hypothesis for the observed robustness in Kangaroo is that the noise induced a higher exploration and reduced overfitting to spurious correlations during training.

\FloatBarrier

\input{sections/llm_interaction}

%% file: sections/logic_policies.tex
\begin{lstlisting}[language=Python, caption={Pong policy of SCoBots \citep{delfosse2024interpretable}}, label={lst:scobots}]
def play(state):
    if state.Ball_0.prev_x <= 1.24:
        if state.Ball_0.y - state.Enemy_0.prev_y  <= -0.41:
            if state.Ball_0.x - state.Ball_0.prev_x  <= 0.09:
                if state.Player_0.y - state.Enemy_0.y  <= -0.91:
                    if state.Ball_0.x <= 0.73:
                        if state.Player_0.y - state.Ball_0.x  <= -1.20:
                            return "LEFT"
                        else:
                            return "RIGHT"
                    else:
                        return "NOOP"
                else:
                    if state.Ball_0.x <= -0.05:
                        return "NOOP"
                    else:
                        if state.Player_0.y - state.Player_0.prev_y  <= -0.19:
                            return "NOOP"
                        else:
                            if state.Ball_0.prev_x - state.Ball_0.prev_y  <= -0.38:
                                return "NOOP"
                            else:
                                if state.Ball_0.x - state.Ball_0.prev_x  <= -0.09:
                                    return "RIGHT"
                                else:
                                    return "RIGHT"
            else:
                return "RIGHT"
        else:
            if state.Ball_0.x - state.Enemy_0.y  <= -1.46:
                if state.Player_0.y - state.Ball_0.prev_x  <= -0.92:
                    return "LEFT"
                else:
                    return "RIGHT"
            else:
                if state.Player_0.y - state.Ball_0.y  <= -0.63:
                    if state.Ball_0.prev_x <= -0.05:
                        return "LEFT"
                    else:
                        if state.Player_0.y - state.Ball_0.y  <= -0.80:
                            return "LEFT"
                        else:
                            if state.Ball_0.prev_x <= 0.80:
                                return "LEFT"
                            else:
                                return "NOOP"
                else:
                    if state.Ball_0.x <= -0.07:
                        if state.Ball_0.prev_y <= -0.15:
                            if state.Ball_0.prev_y <= -1.96:
                                return "LEFT"
                            else:
                                return "LEFT"
                        else:
                            return "NOOP"
                    else:
                        if state.Ball_0.y <= 1.24:
                            if state.Player_0.y - state.Ball_0.y  <= -0.48:
                                if state.Ball_0.prev_x <= 0.68:
                                    return "NOOP"
                                else:
                                    return "NOOP"
                            else:
                                if state.Ball_0.x <= 0.54:
                                    if state.Player_0.y <= -0.82:
                                        return "NOOP"
                                    else:
                                        if state.Enemy_0.y - state.Enemy_0.prev_y  <= -0.09:
                                            return "RIGHT"
                                        else:
                                            return "RIGHT"
                                else:
                                    if state.Ball_0.x - state.Ball_0.y  <= -0.21:
                                        return "NOOP"
                                    else:
                                        return "RIGHT"
                        else:
                            if state.Ball_0.y <= 1.54:
                                return "NOOP"
                            else:
                                return "LEFT"
    else:
        if state.Player_0.y - state.Ball_0.y  <= -0.54:
            if state.Player_0.y - state.Ball_0.y  <= -0.88:
                return "LEFT"
            else:
                if state.Ball_0.y <= -1.40:
                    return "RIGHT"
                else:
                    return "NOOP"
        else:
            if state.Enemy_0.y - state.Enemy_0.prev_y  <= 0.93:
                return "RIGHT"
            else:
                return "LEFT"
\end{lstlisting}

\begin{lstlisting}[caption={Seaquest policy of NUDGE \citep{delfosse2023interpretable}}, label={lst:nudge}]
up_air(X):-oxygen_low(B).
up_divers_collected(X):-all_divers_collected(D).
fire_left(X):-same_depth_enemy(P,E),visible_enemy(E),facing_left(P),right_of_enemy(P,E).
fire_right(X):-same_depth_enemy(P,E),visible_enemy(E),facing_right(P),left_of_enemy(P,E).
left_aim(X):-right_of_enemy(P,E),facing_right(P),same_depth_enemy(P,E),visible_enemy(E).
right_aim(X):-left_of_enemy(P,E),facing_left(P),same_depth_enemy(P,E),visible_enemy(E).
down_aim(X):-higher_than_enemy(P,E),visible_enemy(E).
up_aim(X):-deeper_than_enemy(P,E),visible_enemy(E).
up_evade(X):-close_by_enemy(P,E),same_depth_enemy(P,E),visible_enemy(E).
down_evade(X):-close_by_enemy(P,E),same_depth_enemy(P,E),visible_enemy(E).
up_evade(X):-close_by_missile(P,M),same_depth_missile(P,M),visible_missile(M).
down_evade(X):-close_by_missile(P,M),same_depth_missile(P,M),visible_missile(M).
left_to_diver(X):-right_of_diver(P,D),close_by_diver(P,D),visible_diver(D).
right_to_diver(X):-left_of_diver(P,D),close_by_diver(P,D),visible_diver(D).
up_to_diver(X):-deeper_than_diver(P,D),close_by_diver(P,D),visible_diver(D).
down_to_diver(X):-higher_than_diver(P,D),close_by_diver(P,D),visible_diver(D).
\end{lstlisting}

\begin{lstlisting}[caption={Pong policy of INSIGHT \citep{luo2024endtoend}}, label={lst:INSIGHT}]
logits_noop2 = -0.074*y_agent_1*y_opponent_2 + 0.059*y_agent_1*y_opponent_3 - 0.097*y_agent_4 - 0.16*y_opponent_1*y_opponent_2 - 0.18*y_opponent_2**2 - 0.27*y_opponent_2 + 0.063*y_opponent_4 
logits_up1 = 0.23*y_agent_1**2 + 0.59*y_agent_1*y_agent_2 + 0.4*y_agent_2**2 + 0.11*y_agent_2 - 1.5*y_agent_4**2 - 3.6*y_agent_4 + 0.068*y_opponent_3 + 1.1
logits_down1 = 0.09*x_ball_3 + 0.12*x_ball_4 - 0.21*y_agent_1**2 + 0.12*y_agent_1*y_opponent_1 + 0.27*y_agent_1*y_opponent_2 - 0.43*y_agent_1*y_opponent_3 - 0.28*y_agent_1*y_opponent_4 + 0.13*y_agent_2 + 0.14*y_agent_4**2 + 0.43*y_agent_4 + 0.087*y_ball_3 + 0.15*y_ball_4 + 0.14*y_opponent_1**2 + 0.6*y_opponent_1*y_opponent_2 + 0.61*y_opponent_1 + 0.65*y_opponent_2**2 + 1.1*y_opponent_2 - 0.2*y_opponent_3**2 - 0.26*y_opponent_3*y_opponent_4 - 2.8*y_opponent_3 - 0.085*y_opponent_4**2 - 0.14*y_opponent_4 - 2.3 
logits_up2 = 0.063*x_ball_4 - 0.078*y_agent_1 + 0.18*y_agent_2**2 + 0.52*y_agent_2*y_agent_3 + 0.35*y_agent_2*y_opponent_1 + 0.29*y_agent_2*y_opponent_2 + 0.26*y_agent_2 + 0.38*y_agent_3**2 + 0.51*y_agent_3*y_opponent_1 + 0.42*y_agent_3*y_opponent_2 + 1.6*y_agent_3 - 8.2*y_agent_4 - 0.085*y_ball_3 + 0.17*y_opponent_1**2 + 0.28*y_opponent_1*y_opponent_2 + 0.25*y_opponent_1 + 0.11*y_opponent_2**2 + 0.15*y_opponent_2 - 0.074*y_opponent_3 + 0.26 logits_down2 = -0.052*x_ball_1 - 0.068*x_ball_3 - 0.093*x_ball_4 + 0.18*y_agent_1 - 0.17*y_agent_2**2 - 0.49*y_agent_2*y_agent_3 - 0.33*y_agent_2*y_opponent_1 - 0.27*y_agent_2*y_opponent_2 - 0.39*y_agent_2 - 0.35*y_agent_3**2 - 0.48*y_agent_3*y_opponent_1 - 0.4*y_agent_3*y_opponent_2 - 0.38*y_agent_3 + 0.15*y_agent_4**2 + 0.54*y_agent_4 - 0.06*y_ball_1 - 0.064*y_ball_3 - 0.11*y_ball_4 - 0.17*y_opponent_1**2 - 0.28*y_opponent_1*y_opponent_2 - 0.58*y_opponent_1 - 0.13*y_opponent_2**2 - 0.38*y_opponent_2 + 2.2*y_opponent_3 - 0.052*y_opponent_4 - 3.6

action_noop = [exp(logits_noop1) + exp(logits_noop2)] / sum(exp(logits)) 
action_up = [exp(logits_up1) + exp(logits_up2)] / sum(exp(logits)) 
action_down = [exp(logits_down1) + exp(logits_down2)] / sum(exp(logits))
\end{lstlisting}

%% file: sections/baseline_comp_table.tex
\begin{table}
\centering
\caption{Performance Comparison on Seaquest and Kangaroo}
\label{tab:performance_comp}
\resizebox{\textwidth}{!}{
\begin{tabular}{l | cc cc || cc cc}
\hline
\multirow{2}{*}{\textbf{Algorithm}} & \multicolumn{4}{c||}{\textbf{Seaquest}} & \multicolumn{4}{c}{\textbf{Kangaroo}} \\
 & Default & Divers Collected & Simplification & Divers Collected & Default & Level Completion (\%) & Simplification & Level Completion (\%) \\
\hline

PPO & 915.3 $\pm$ 101.1 & (6.7 $\pm$ 0.1) & 0.0 $\pm$ 0.0 & (4.7 $\pm$ 0.6) & 550.0 $\pm$ 353.6 & (33.3 $\pm$ 0.0) & 150.0 $\pm$ 212.1 & (33.3 $\pm$ 0.0) \\
PQN & 1361.3 $\pm$ 294.6 & (4.8 $\pm$ 2.8) & 0.0 $\pm$ 0.0 & (4.0 $\pm$ 1.5) & 949.7 $\pm$ 279.9 & (47.3 $\pm$ 0.4) & 338.9 $\pm$ 209.1 & (40.5 $\pm$ 13.6) \\
NEXUS (neural) & 2100.7 $\pm$ 614.5 & (9.2 $\pm$ 4.2) & 0.0 $\pm$ 0.0 & (6.7 $\pm$ 6.8) & 1889.0 $\pm$ 1770.6 & (70.6 $\pm$ 28.8) & 551.0 $\pm$ 513.2 & (58.9 $\pm$ 30.9) \\
NEXUS (symbolic) & \textbf{2229.1 $\pm$ 172.9} & (12.4 $\pm$ 0.7) & 9416.7 $\pm$ 4363.2 & (41.0 $\pm$ 13.1) & 1030.3 $\pm$ 553.3 & (83.8 $\pm$ 2.6) & 448.7 $\pm$ 147.4 & (52.0 $\pm$ 9.4) \\
NEXUS (nesy) & 2185.6 $\pm$ 1000.1 & (10.8 $\pm$ 2.9) & 2483.3 $\pm$ 1607.7 & (15.3 $\pm$ 9.0) & 909.4 $\pm$ 519.1 & (85.6 $\pm$ 12.8) & 674.8 $\pm$ 115.8 & (72.1 $\pm$ 16.9) \\
NUDGE & 46.7 $\pm$ 23.1 & (4.1 $\pm$ 0.7) & 113.3 $\pm$ 99.6 & (7.3 $\pm$ 4.5) & 1522.2 $\pm$ 661.9 & \textbf{(110.0 $\pm$ 17.3)} & \textbf{1966.7 $\pm$ 290.6} & (101.7 $\pm$ 2.9) \\
BlendRL & 2138.9 $\pm$ 1636.0 & \textbf{(15.8 $\pm$ 8.6)} & \textbf{24755.6 $\pm$ 29438.4} & \textbf{(55.6 $\pm$ 43.0)} & \textbf{1955.6 $\pm$ 887.8} & (101.7 $\pm$ 2.9) & 1944.4 $\pm$ 300.6 & \textbf{(118.3 $\pm$ 27.5)} \\
SCoBots & 1055.3 $\pm$ 272.6 & (- $\pm$ -) & 0.0 $\pm$ 0.0 & (- $\pm$ -) & 2776.6 $\pm$ 1332.4 & (- $\pm$ -) & 0.0 $\pm$ 0.0 & (- $\pm$ -) \\
\hline
\end{tabular}
}
\end{table}

%% file: sections/llm_interaction.tex
\section{LLM Interaction}
\label{sec:llm_interaction}
We outline the procedure for leveraging an LLM to generate task-relevant skills, associated reward functions, and a fixed meta-policy rule set. The LLM is conditioned on the game's original manual and structured type information describing the object-centric observations available to the agent.

Initially, the LLM is queried for a set of skills and corresponding reward functions. These outputs can be manually refined before querying the LLM for a meta-policy function that selects which skill to execute. Prior prompts and responses are retained to maintain conversational context, consistent with standard chat behavior. We provide the entire prompt for the game Kangaroo in \Cref{prompts}. The prompt for Seaquest was generated equivalently.

Unedited responses from GPT-4o (via chatgpt.com on 21-07-2025) for the games Kangaroo and Seaquest are included in \Cref{gen:1}–\Cref{gen:4}.

For Crafter, we used the LLM to generate the list of important tasks, as well as the symbolic meta-policy function. Prompt and answers are available in \Cref{gen:5} and \Cref{gen:6}. Rewards for the specific skills were crafted manually.

The LLM outputs served as a starting point: while the proposed skill decompositions and the structure of the meta-policies were largely retained, the final implementations required both framework adaptations (e.g., JAX compatibility, mapping to the available state variables) and semantic refinements by the authors, such as setting distance and resource thresholds, reordering rule priorities, and simplifying or rebalancing reward terms. The neuro-symbolic rule sets were derived manually from the symbolic meta-policies by making their conditions non-exclusive. \Cref{tab:llm_edits} lists all changes per environment and component; the final code is available in our repository\footnote{\url{https://github.com/remunds/symbolic_options}}.

\newcommand{\placeholder}[1]{\texttt{<#1>}}

\newtcolorbox{promptbox}{
  colback=gray!5,
  colframe=gray!80,
  boxrule=0.5pt,
  arc=2pt,
  left=4pt,
  right=4pt,
  top=4pt,
  bottom=4pt,
  fontupper=\small,
  width=\linewidth
}

\begin{figure}[tb]
\centering

\begin{promptbox}
You are a RL expert and develop a hierarchical agent to play the atari game Seaquest. Below, I have provided a detailed description of the game. Your task is to come up with a short list of essential skills (3-4) that are needed to solve the game.

Game Description:\\
---\\
\placeholder{Atari game manual\footnote{copied entirely from \url{https://www.atariage.com/manual_html_page.php?SoftwareLabelID=923}}}\\
---

Think about what the main goal of this game is, then come up with a list of the required skills to solve the game and provide it in the output.
Finally, create a pseudo reward function for each of these skills that a RL agent can use to learn the skills.

The reward functions are called at each step in the environment and retrieve the current and previous symbolic observation of the game.
Here is the type definition for the observation:\\
---
\begin{lstlisting}[language=Python]
class KangarooObservation(NamedTuple):
    player_x: chex.Array
    player_y: chex.Array
    player_o: chex.Array
    platform_positions: chex.Array
    ladder_positions: chex.Array
    fruit_positions: chex.Array
    bell_position: chex.Array
    child_position: chex.Array
    falling_coco_position: chex.Array
    monkey_positions: chex.Array
    coco_positions: chex.Array
\end{lstlisting}
---\\
Each skill and it's reward function should be as simple as possible that achieves the wanted behavior.\\
Provide: List of skills and description, pseudo-code reward functions for each skill.
\end{promptbox}

\vspace{1em}

\begin{promptbox}
You are now creating a function that decides at each time step in the game, which skill to choose.
The skills are\footnote{copied from answer to previous prompt, see \Cref{gen:2}.}:
\begin{itemize}
    \item Obstacle Avoidance
    \item Vertical Navigation
    \item Fruit Collection
    \item Goal Reaching (Child Rescue)
\end{itemize}
The function again has access to the current and previous observation.

Think about at what point in the game you want to use which skill, then provide the function.
\end{promptbox}
\caption{{Subsequent prompts (chat-function) to generate a skill-list and corresponding reward functions (top) and meta-policy function (bottom) for the game Kangaroo.}}
\label{prompts}
\end{figure}

\begin{figure*}[tb]
\centering
\begin{tcolorbox}[title=Essential Skills for Solving \textit{Kangaroo} (Atari 2600), colback=gray!3, colframe=black, width=\textwidth, sharp corners=south]

A hierarchical RL agent for \textit{Kangaroo} can be built upon the following four primitive skills:

\vspace{0.5em}
\textbf{1. Obstacle Avoidance} \\
\textit{Description:} Avoid contact with apples or monkeys via dodging or punching.

\textit{Reward Function:}
\begin{lstlisting}[language=Python]
def reward_obstacle_avoidance(obs, prev_obs):
    collided = any(np.array_equal(obs.player_x, x) and 
                   np.array_equal(obs.player_y, y)
                   for (x, y) in obs.apple_positions + obs.monkey_positions)
    return -1.0 if collided else 0.0
\end{lstlisting}

\vspace{0.5em}
\textbf{2. Vertical Navigation} \\
\textit{Description:} Climb ladders to ascend levels and progress.

\textit{Reward Function:}
\begin{lstlisting}[language=Python]
def reward_vertical_navigation(obs, prev_obs):
    dy = prev_obs.player_y - obs.player_y
    return 0.1 if dy > 0 else 0.0
\end{lstlisting}

\vspace{0.5em}
\textbf{3. Fruit Collection} \\
\textit{Description:} Collect fruits and optionally ring a bell to regenerate them.

\textit{Reward Function:}
\begin{lstlisting}[language=Python]
def reward_fruit_collection(obs, prev_obs):
    reward = 0.0
    if len(obs.fruit_positions) < len(prev_obs.fruit_positions):
        reward += 0.5
    if np.array_equal(obs.player_x, obs.bell_position[0]) and \
       np.array_equal(obs.player_y, obs.bell_position[1]):
        reward += 0.2
    return reward
\end{lstlisting}

\vspace{0.5em}
\textbf{4. Goal Reaching (Child Rescue)} \\
\textit{Description:} Reach the child's location to complete the level.

\textit{Reward Function:}
\begin{lstlisting}[language=Python]
def reward_reach_child(obs, prev_obs):
    return 1.0 if np.array_equal(obs.player_x, obs.child_position[0]) and \
                  np.array_equal(obs.player_y, obs.child_position[1]) else 0.0
\end{lstlisting}

\vspace{0.5em}
These modular rewards enable a high-level controller to invoke appropriate behaviors depending on context.
\end{tcolorbox}
\caption{Generated reward functions for Kangaroo by GPT4-O}
\label{gen:1}
\end{figure*}

\begin{figure*}[tb]
\centering
\begin{tcolorbox}[title=Kangaroo Skill Selection Function, colback=gray!3, colframe=black, width=\textwidth, sharp corners=south]

To select appropriate skills, the agent follows a fixed priority:

\begin{enumerate}
    \item \textbf{Obstacle Avoidance}: If any apple or monkey is in close proximity.
    \item \textbf{Goal Reaching}: If the player is near the child on the top platform.
    \item \textbf{Fruit Collection}: If fruit is visible and reachable.
    \item \textbf{Vertical Navigation}: Default behavior to explore or ascend.
\end{enumerate}

\vspace{0.5em}
\textbf{Skill Selector Function:}
\begin{lstlisting}[language=Python]
def select_skill(obs, prev_obs):
    player_pos = np.array([obs.player_x, obs.player_y])
    
    def in_danger():
        danger_entities = obs.apple_positions + obs.monkey_positions
        return any(np.linalg.norm(player_pos - np.array([x, y])) <= 1.0
                   for (x, y) in danger_entities)

    def near_child():
        child_pos = np.array([obs.child_position[0], obs.child_position[1]])
        return np.linalg.norm(player_pos - child_pos) <= 1.5

    def fruit_visible():
        return any(abs(y - obs.player_y) <= 1 and 
                   abs(x - obs.player_x) <= 3 
                   for (x, y) in obs.fruit_positions)

    if in_danger():
        return obstacle_avoidance
    elif near_child():
        return goal_reaching
    elif fruit_visible():
        return fruit_collection
    else:
        return vertical_navigation
\end{lstlisting}

\vspace{0.5em}
This rule-based selection policy balances safety, progression, and reward acquisition, enabling effective hierarchical control.
\end{tcolorbox}
\caption{Generated fixed meta-policy function for Kangaroo by GPT4-O}
\label{gen:2}
\end{figure*}

\begin{figure*}[tb]
\centering
\begin{tcolorbox}[title=Essential Skills for Solving \textit{Seaquest} (Atari 2600), colback=gray!3, colframe=black, width=\textwidth, sharp corners=south]
To solve \textit{Seaquest} effectively, a hierarchical RL agent must master the following four core skills:

\vspace{0.5em}
\textbf{1. Avoid Obstacles (Survival Navigation)} \\
\textit{Description:} Avoid collisions with sharks, enemy submarines, missiles, and the patrol sub, both underwater and when surfacing.

\textit{Reward Function (pseudo-code):}
\begin{lstlisting}[language=Python]
def avoid_obstacles_reward(prev_obs, curr_obs):
    if player_collided(curr_obs):
        return -1.0
    return 0.0

def player_collided(obs):
    return check_collision_with_sharks(obs) or \
           check_collision_with_subs(obs) or \
           check_collision_with_missiles(obs) or \
           check_collision_with_surface_sub(obs)
\end{lstlisting}

\vspace{0.5em}
\textbf{2. Collect Divers} \\
\textit{Description:} Navigate to and collect divers to accumulate score and progress.

\textit{Reward Function (pseudo-code):}
\begin{lstlisting}[language=Python]
def collect_divers_reward(prev_obs, curr_obs):
    prev_count = count_nearby_divers(prev_obs)
    curr_count = count_nearby_divers(curr_obs)
    return float(prev_count > curr_count) * 1.0

def count_nearby_divers(obs):
    return sum(is_near(obs.player_x, obs.player_y, dx, dy)
               for dx, dy in obs.diver_positions)
\end{lstlisting}

\vspace{0.5em}
\textbf{3. Surface Strategically} \\
\textit{Description:} Surface safely when carrying six divers or when oxygen is low, to drop off divers and replenish oxygen.

\textit{Reward Function (pseudo-code):}
\begin{lstlisting}[language=Python]
def surface_reward(prev_obs, curr_obs):
    surfaced_now = prev_obs.player_y > 0 and curr_obs.player_y == 0
    has_6_divers = num_divers_collected(prev_obs) == 6
    low_oxygen = prev_obs.oxygen < OXYGEN_THRESHOLD
    if surfaced_now and (has_6_divers or low_oxygen):
        return 2.0 if has_6_divers else 0.5
    return 0.0

def num_divers_collected(obs):
    return 6 - sum(is_valid(dx, dy) for dx, dy in obs.diver_positions)
\end{lstlisting}

\vspace{0.5em}
\textbf{4. Eliminate Threats (Combat)} \\
\textit{Description:} Fire missiles to destroy sharks and enemy submarines, clearing paths and increasing score.

\textit{Reward Function (pseudo-code):}
\begin{lstlisting}[language=Python]
def eliminate_threats_reward(prev_obs, curr_obs):
    num_enemies_prev = count_enemies(prev_obs)
    num_enemies_curr = count_enemies(curr_obs)
    return float(num_enemies_curr < num_enemies_prev) * 0.5

def count_enemies(obs):
    return sum(is_valid(dx, dy) for dx, dy in obs.shark_positions) + \
           sum(is_valid(dx, dy) for dx, dy in obs.sub_positions)
\end{lstlisting}

\end{tcolorbox}
\caption{Generated reward functions for Seaquest by GPT4-O}
\label{gen:3}
\end{figure*}

\begin{figure*}[tb]
\centering
\begin{tcolorbox}[title=Seaquest Skill Selection Function, colback=gray!3, colframe=black, width=\textwidth, sharp corners=south]

Here is a skill selection function that chooses the appropriate control skill at each time step based on game context.

\vspace{0.5em}
\textbf{Skill Selection Function (pseudo-code):}
\begin{lstlisting}[language=Python]
def select_skill(prev_obs, curr_obs):
    # Skill names
    AVOID = "obstacle_avoidance"
    COLLECT = "collect_divers"
    SURFACE = "surface_strategically"
    COMBAT = "eliminate_threats"

    # === Utility Functions ===
    def is_threat_near(obs: SeaquestObservation) -> bool:
        return check_collision_with_sharks(obs) or \
               check_collision_with_subs(obs) or \
               check_collision_with_missiles(obs) or \
               check_collision_with_surface_sub(obs)

    def num_divers_collected(obs: SeaquestObservation) -> int:
        return 6 - sum(is_valid(x, y) for x, y in obs.diver_positions)

    def low_oxygen(obs: SeaquestObservation) -> bool:
        return obs.oxygen < 20  # adjustable threshold

    def at_surface(obs: SeaquestObservation) -> bool:
        return obs.player_y == 0

    # === Skill Selection Logic ===
    if is_threat_near(curr_obs):
        return AVOID

    if (num_divers_collected(curr_obs) == 6 or low_oxygen(curr_obs)) and \
        not at_surface(curr_obs):
        return SURFACE

    if has_nearby_enemies(curr_obs):  # e.g. visible on screen in front
        return COMBAT

    return COLLECT
\end{lstlisting}

\vspace{0.5em}
\textbf{Explanation of Logic:}
\begin{itemize}
    \item \textbf{Obstacle Avoidance} is triggered immediately when any nearby threat is detected.
    \item \textbf{Surface Strategically} is invoked if the agent has collected 6 divers or is low on oxygen, and is not already at the surface.
    \item \textbf{Eliminate Threats} is used when enemies are visible but not an immediate threat.
    \item \textbf{Collect Divers} is the default skill when no critical conditions are active.
\end{itemize}

This rule-based selection mechanism allows a high-level controller to choose among primitive skills in a safety-first manner while still enabling reward-driven exploration and progress.
\end{tcolorbox}
\caption{Generated fixed meta-policy for Seaquest by GPT4-O}
\label{gen:4}
\end{figure*}

\begin{figure}[tb]
\centering
\begin{promptbox}
You are a reinforcement learning specialist. The following is the basic description of the game Crafter, which you are trying to solve.

----

Crafter is a game about exploring dungeons, mining, crafting and fighting enemies. The player can move in the four cardinal directions using WASD and can interact using SPACE. Interacting can cause the player to attempt to mine (a block), attack (a creature), drink (water or from a fountain), or eat (fruit).

The player has the 'intrinsics' health, hunger, thirst and energy. Hunger, thirst and energy will naturally decrease and must be replenished by eating, drinking and sleeping respectively. Health will recover when hunger, thirst and energy are non-zero and will decrease if any of these are 0. If the players health falls beneath 0 they will die and the game will restart.

----

First, come up with 5 basic skills that can be used to solve the game. Only provide necessary skills without duplications.
After that, please come up with a simple function (pseudo code) that gets the game-state as input and returns an integer deciding on the active skill to select in the current state. The skills correspond to the ones that you chose before.
Try to solve the game as far as possible.

The game-state looks like this: 
\begin{lstlisting}[language=Python]
@struct.dataclass
class Inventory:
    wood: int = 0
    stone: int = 0
    coal: int = 0
    iron: int = 0
    diamond: int = 0
    sapling: int = 0
    wood_pickaxe: int = 0
    stone_pickaxe: int = 0
    iron_pickaxe: int = 0
    wood_sword: int = 0
    stone_sword: int = 0
    iron_sword: int = 0

@struct.dataclass
class Mobs:
    position: jnp.ndarray
    health: int
    mask: bool
    attack_cooldown: int

@struct.dataclass
class EnvState:
    map: jnp.ndarray
    mob_map: jnp.ndarray
    player_position: jnp.ndarray
    player_direction: int
    # Intrinsics
    player_health: int
    player_food: int
    player_drink: int
    player_energy: int
    is_sleeping: bool
    # Second order intrinsics
    player_recover: float
    player_hunger: float
    player_thirst: float
    player_fatigue: float
    inventory: Inventory
    zombies: Mobs
    cows: Mobs
    skeletons: Mobs
    arrows: Mobs
    arrow_directions: jnp.ndarray
    growing_plants_positions: jnp.ndarray
    growing_plants_age: jnp.ndarray
    growing_plants_mask: jnp.ndarray
\end{lstlisting}
\end{promptbox}
\caption{Prompt used to generate a skill-list and meta-policy function for Crafter.}
\label{gen:5}
\end{figure}

\begin{figure*}[tb]
\centering
\begin{tcolorbox}[title=Essential Skills for Solving \textit{Crafter}, colback=gray!3, colframe=black, width=\textwidth, sharp corners=south]

\textbf{Basic Skills.}  
We identify the following five skills as sufficient building blocks for solving \textit{Crafter}:  

\begin{enumerate}
    \item \textbf{Gather Resources:} Mine wood, stone, coal, iron, or other materials when available.  
    \item \textbf{Craft Tools/Weapons:} Use gathered resources to craft pickaxes and swords for efficiency and combat readiness.  
    \item \textbf{Eat/Drink/Sleep:} Replenish food, water, or energy to maintain survival.  
    \item \textbf{Fight/Defend:} Attack hostile mobs when they are nearby and health is sufficient.  
    \item \textbf{Explore/Move:} Traverse unexplored areas to discover resources, water, and safe locations.  
\end{enumerate}

\medskip
\textbf{Policy Function.}  
The following pseudo-code outlines a rule-based strategy for selecting the active skill based on the current game state.  

\begin{lstlisting}[language=Python, basicstyle=\ttfamily\footnotesize, frame=single]
def select_skill(state: EnvState) -> int:
    """
    Returns the skill index:
    0 = Gather Resources
    1 = Craft Tools/Weapons
    2 = Eat/Drink/Sleep
    3 = Fight/Defend
    4 = Explore/Move
    """

    # --- Survival priority ---
    if state.player_food <= 2 or state.player_drink <= 2 or state.player_energy <= 2:
        return 2  # Eat/Drink/Sleep

    # --- Combat ---
    if mob_nearby(state) and state.player_health > 3:
        return 3  # Fight/Defend

    # --- Resource management ---
    if resources_nearby(state) and inventory_needs_tools(state.inventory):
        return 0  # Gather Resources

    if can_craft_better_tools(state.inventory):
        return 1  # Craft Tools/Weapons

    # --- Default action ---
    return 4
\end{lstlisting}

\end{tcolorbox}
\caption{Generated list of essential skills and a rule-based policy for solving \textit{Crafter} by GPT4-O. The policy prioritizes survival, then combat, followed by resource gathering and crafting, and defaults to exploration otherwise.}
\label{gen:6}
\end{figure*}

\begin{table}[t]
\centering\small
\caption{Human edits to LLM-generated components (GPT-4o, App.~\ref{sec:llm_interaction}). F: framework adaptation (JAX, state fields); S: semantic change; M: written manually.}
\label{tab:llm_edits}
\begin{tabularx}{\linewidth}{@{}llXc@{}}
\toprule
Env. & Component & Change from LLM output to final implementation & Type \\
\midrule
\multirow{5}{*}{Seaquest}
 & Skill set & 4 skills $\rightarrow$ 3; \emph{avoid obstacles} dropped (death penalty moved into \emph{fight}) & S \\
 & $r_\text{collect}$ & diver proximity count $\rightarrow$ direct \texttt{divers\_collected} counter & F \\
 & $r_\text{fight}$ & enemy count decrease (+0.5) $\rightarrow$ score increase (+1), life lost ($-1$) & S \\
 & $r_\text{surface}$ & surfacing with 6 divers / low oxygen $\rightarrow$ oxygen refill (+0.1) & S \\
 & Meta-policy & same conditions; thresholds set (enemy $<$40px, oxygen $<$10), priority reordered & F+S \\
\midrule
\multirow{5}{*}{Kangaroo}
 & Skill set & 4 skills $\rightarrow$ 3; \emph{goal reaching} merged into \emph{navigate} & S \\
 & $r_\text{navigate}$ & +0.1 per upward step $\rightarrow$ clipped $\Delta y$, penalty below floor 1, 0 when crashing & S \\
 & $r_\text{enemies}$ & collision ($-1$) $\rightarrow$ monkey killed (+1), life lost ($-1$) & S \\
 & $r_\text{fruits}$ & fruit (+0.5) and bell (+0.2) $\rightarrow$ fruit (+1) & S \\
 & Meta-policy & same structure without goal rule; distance thresholds set (35/50px) & F+S \\
\midrule
\multirow{3}{*}{Crafter}
 & Skill set & 5 skills kept (renamed) & -- \\
 & Rewards & written manually & M \\
 & Meta-policy & helper predicates implemented; thresholds changed; combat before survival; crafting before collection & F+S \\
\midrule
All & NeSy rules & derived manually from the symbolic meta-policy by making conditions non-exclusive & M \\
\bottomrule
\end{tabularx}
\end{table}